%% file: lm_reasoning_acl.tex
\pdfoutput=1

\documentclass[11pt]{article}

\usepackage[final]{acl}

\usepackage{times}
\usepackage{latexsym}

\usepackage[T1]{fontenc}

\usepackage[utf8]{inputenc}

\usepackage{microtype}

\usepackage{inconsolata}

\usepackage{hyperref}
\usepackage{url}
\usepackage{graphicx}
\usepackage{booktabs}
\usepackage{lipsum} 
\usepackage{amssymb}
\usepackage{comment}
\usepackage{afterpage}
\usepackage{ragged2e} 
\usepackage{color}
\usepackage{adjustbox}
\usepackage{float}
\usepackage{enumitem}
\usepackage{wrapfig}
\newcommand{\hide}[1]{}
\usepackage{fancybox}
\usepackage{subcaption}
\usepackage{array}

\input{math_commands.tex}

\usepackage{tcolorbox}

\usepackage{tikz}

\newcounter{tcolorboxcounter}

\title{A Systematic Comparison of Syllogistic\\Reasoning in Humans and Language
Models}

\newcommand{\blfootnote}[1]{%
\begingroup
\renewcommand{\thefootnote}{}
\footnote{#1}%
\addtocounter{footnote}{-1}%
\endgroup }

\author{\textbf{Tiwalayo Eisape,$^{\dagger}$}
\textbf{MH Tessler,$^{\ddagger}$}
\textbf{Ishita Dasgupta,$^{\ddagger}$}
\textbf{Fei Sha,$^{\S}$}\\
\textbf{Sjoerd van Steenkiste,$^{\S,*}$}
\textbf{Tal Linzen$^{\S,*}$}\\ Massachusetts Institute of Technology$^{\dagger}$,
Google DeepMind$^{\ddagger}$, Google Research$^{\S}$ }

\begin{document}
  \maketitle

  \begin{abstract}
A central component of rational behavior is logical inference: the process
of determining which conclusions follow from a set of premises. Psychologists
have documented several ways in which humans' inferences deviate from the
rules of logic. Do language models, which are trained on text generated by
humans, replicate such human biases, or are they able to overcome them? Focusing
on the case of syllogisms---inferences from two simple premises---we show that, within the PaLM~2 family of transformer language models,
larger models are more logical than smaller ones, and also more logical than
humans. At the same time, even the largest models make systematic errors, some
of which mirror human reasoning biases: they show sensitivity to the (irrelevant)
ordering of the variables in the syllogism, and draw confident but incorrect inferences from particular
syllogisms (syllogistic fallacies). Overall, we find that language models often
mimic the human biases included in their training data, but are able to
overcome them in some cases.
  \end{abstract}
  \blfootnote{\kern-1.9em$^{\dagger}$Work done when TE was a student researcher at Google Research. $^{*}$SVS and TL are joint senior authors. Author contributions: TE, SVS, and TL co-led the project. TE conducted the experiments and analysis. TE, MHT, ID, FS, SVS, and TL helped with project framing, analysis and suggesting experiments. ID and SVS offered technical guidance and help with engineering. TE wrote the paper with help from MHT, ID, FS, SVS, and TL. Correspondence: \href{mailto:eisape@mit.edu}{\texttt{eisape@mit.edu}}, \href{mailto:svansteenkiste@google.com}{\texttt{svansteenkiste@google.com}}, \href{mailto:linzen@google.com}{\texttt{linzen@google.com}}.}

  \input{2_intro}
  \input{3_background}
  \input{4_method}

\input{5_results}
  \input{6_mreasoner}
  \input{7_discuss}

  \bibliography{lm_reasoning_iclr}

  \appendix
  \input{8_appendix}
\end{document}

%% file: math_commands.tex
\usepackage{amsmath,amsfonts,bm}

\def\eqref#1{equation~\ref{#1}}

\def\1{\bm{1}}

\DeclareMathAlphabet{\mathsfit}{\encodingdefault}{\sfdefault}{m}{sl}
\SetMathAlphabet{\mathsfit}{bold}{\encodingdefault}{\sfdefault}{bx}{n}



%% file: 2_intro.tex
\section{Introduction}
\label{sIntro} The capacity to reason deductively---that is, to determine which inferences,
if any, follow from a given set of premises---is central to rational thought \citep{Newell1972-tr,Laird1987-so,Fodor1988-qw,Griffiths2010-ff}.
Despite the importance of this capacity, human reasoning often displays
systematic biases \citep{Gigerenzer2011-as,Marcus2009-lz,Kahneman2013-wd,McClelland2010-ce}.
In recent years, language models (LMs) trained with self-supervised objectives have
been reported to display a range of capabilities, including the ability to
reason \citep{brown2020language,chowdhery2022palm,Bubeck2023-pn}. Does LMs' logical
reasoning follow the rules of logic to a greater extent than humans'? To the extent
that LMs' reasoning deviates from normative logic, are their biases similar to humans'
\citep{binz2023using,Dasgupta2022-ln}?

In this work, we address these questions with a detailed study of a particularly
simple case---inferences from pairs of premises, or \textit{syllogisms}, such as
the following: \\
\begin{center}
    \begin{minipage}{0.47\textwidth}
        \begin{tcolorbox}
            [colback=gray!5!white,colframe=black!75!black, center upper, center
            lower]
            \stepcounter{tcolorboxcounter}
            \label{example:EA1}
            If \textbf{all bakers are artists}, \\
            and \textbf{some bakers are chemists},
            \tcblower
            then: \textbf{some artists are chemists}.
        \end{tcolorbox}
    \end{minipage}
\end{center}
\begin{table*}
    \centering
    \begin{minipage}[b]{0.5\linewidth}
        \centering
        \begin{tabular}{clcl}
            \textbf{A}: & \textbf{All} artists are bakers & \textbf{I}: & \textbf{Some} artists are bakers              \\
            \textbf{E}: & \textbf{No} artists are bakers  & \textbf{O}: & \textbf{Some} artists \textbf{are not} bakers \\
        \end{tabular}
    \end{minipage}%
    \hfill 
    \begin{minipage}[b]{0.4\linewidth}
        \centering
        \begin{tabular}{cccc}
            1            & 2   & 3   & 4   \\
            \midrule A-B & B-A & A-B & B-A \\
            B-C          & C-B & C-B & B-C \\
        \end{tabular}
    \end{minipage}
    \caption{Syllogism moods (left) and variable orderings (right).\label{table:moods-variables}}
\end{table*}
\vspace{1em}
In a syllogism, each premise relates two terms with one of four quantifiers (traditionally
known as ``moods''): \emph{all}, \emph{some}, \emph{none} and \emph{some are not}.
Only one term is shared between the premises (\emph{bakers} in the example above). Inference is required to determine if there is a necessary relationship between
the two remaining terms (here, \emph{artists} and \emph{chemists}) when the
premises in question are true.

When human participants in experiments are asked to make syllogistic inferences, their responses often deviate
from the rules of logic; in fact, for some syllogisms the vast majority of participants
draw incorrect inferences \citep{Khemlani2012-mj}. This could pose a challenge to
language models (LMs), as they learn from corpora consisting primarily of human-generated
texts---texts which, in turn, reflect human beliefs and inferences. Is there sufficient signal in
the training corpus to steer LMs away from (often incorrect) human
inferences and toward a behavior consistent with the normative rules of logic---the behavior that is desirable
for most applications?

We address this question in a detailed comparison between the PaLM 2 family of transformer
LMs~\citep{google2023palm2} and studies from cognitive psychology, as well as a replication with the Llama~2  family of transformer LMs \cite{touvron2023llama}. We report the
following results:
\begin{enumerate}[leftmargin=*]
    \item LMs draw correct inferences more often than humans, and larger LMs tend to be more accurate than smaller ones, but the accuracy of even the
        best performing LM is only about 75\%, and scale does not consistently
        lead to accuracy gains (Section~\ref{sec:lms-logical}).

    \item LM errors are systematic, with very low accuracy on particular
        syllogism types (Section~\ref{sec:lms-logical}); the syllogisms that LMs
        struggle with are a subset of those that humans find difficult (Section~\ref{sec:lms-humanlike}).

    \item Like humans, LMs are sensitive to the ordering of terms in the
        premises of a syllogism even when it is logically irrelevant (Section~\ref{sec:figural-effects};
        this pattern is known as the ``figural effect'' in cognitive psychology;
        \citealt{Johnson-Laird1978-jr}).

    \item LMs show many of the same \textit{syllogistic fallacies}, characterized
        by high confidence and low accuracy, as humans. Larger LMs are somewhat more
        susceptible to these fallacies than smaller ones (Section~\ref{sec:fallacies};
        \citealt{Khemlani2017-zi}).

    \item Using the Mental Models theory from cognitive psychology, we find
        quantitative evidence that larger LMs reason more deliberatively than
        smaller ones (Section~\ref{sec:mentalmodels}; \citealt{Khemlani2022-yw}).
\end{enumerate}

Overall, we find that PaLM~2 LMs replicate many of the human biases discovered in
psychology studies, consistent with the fact that LMs are trained on human-generated
text. For some syllogisms, however, sufficiently large models overcome those biases and achieve
dramatically better accuracy than humans, although their overall accuracy
is still far from the perfect logical reasoner.

%% file: 3_background.tex
\section{Background and Related Work}
\subsection{Syllogisms}
\label{sec:syllogisms}
Syllogisms are logical arguments consisting of two \textit{premises} relating
three variables, A, B and C (e.g., \emph{artists}, \emph{bakers}, and \emph{chemists}
in the example from the introduction). Each premise relates just two of the
variables, through one of four quantifiers,
often referred to as ``moods'' (Table~\ref{table:moods-variables}, left). The
variables in each of the premises can be ordered in either of the two directions---e.g.,
\emph{all artists are bakers} vs. \emph{all bakers are artists}---and so there are
four possible pairs of orderings (Table~\ref{table:moods-variables}, right). These
orderings are traditionally referred to as ``figures'', but we will use the more
transparent term ``variable ordering''. Taking the cross product of these building
blocks yields 64 possible syllogisms: two premises, each of which can take one of
four quantifiers and one of two possible orderings.

Though the premises only relate A and B, or B and C---never A and C---27 of the
64 syllogisms imply a quantified relationship between A and C (e.g., \emph{some
A are C}). In the remaining 37 syllogisms, no relation between A and C can be deduced;
in human experiments, the expected response to these syllogisms is ``nothing follows'' (see Figure~\ref{fig:valid-conclusions} in Appendix~\ref{sec:content-words} for the full set of valid conclusions for each syllogism).

\subsection{Human Syllogistic Reasoning}
Psychologists, going back to the early 20th century, have found that the conclusions
that humans draw from the premises of a syllogism often deviate from logical norms
(for a review, see \citealt{Khemlani2012-mj}). These errors are systematic: some
syllogisms are much harder than others, and the incorrect conclusions that
participants tend to draw are consistent across participants. For example, from the two premises (1) \emph{no artists are bakers} and (2) \emph{all bakers are chemists}, the
vast majority of participants incorrectly conclude that it is the case that \emph{no artists are chemists}. We analyse such cases in detail
in Section~\ref{sec:fallacies}.

In addition to these specific, highly challenging syllogisms, several broader
reasoning biases have been documented. When given a syllogistic argument where the
variables in the premises are ordered ``A-B, B-C'', participants show a bias towards
conclusions with an A-C ordering, even though reordering the variables in the premises
does not affect the conclusions licensed by the syllogism \citep{Johnson-Laird1978-jr}. Participants
are also more likely to produce a conclusion when it is true in the real world,
independently of whether it follows from the premises (``content effects'',
\citealt{Evans1983-sm}).

A number of theories have been proposed to explain human syllogistic reasoning. An
influential account that we focus on in this work is the Mental Models Theory \citep{johnson1991deduction}.
This theory posits that human reasoners construct mental models populated by a small
number of entities that instantiate the premises; e.g., to instantiate the
premise \emph{all artists are bakers}, a reasoner might construct a world with
three specific artists, all of whom are bakers. These worlds are constructed
based on a number of fallible heuristics, and human reasoning errors arise when those
heuristics produce incorrect conclusions (see Section~\ref{sec:mentalmodels}).

\subsection{Language Models and Reasoning}
\begin{figure}
    \centering
    \includegraphics[width=.7\linewidth, clip, trim=25cm 5cm 25cm 5cm]{
        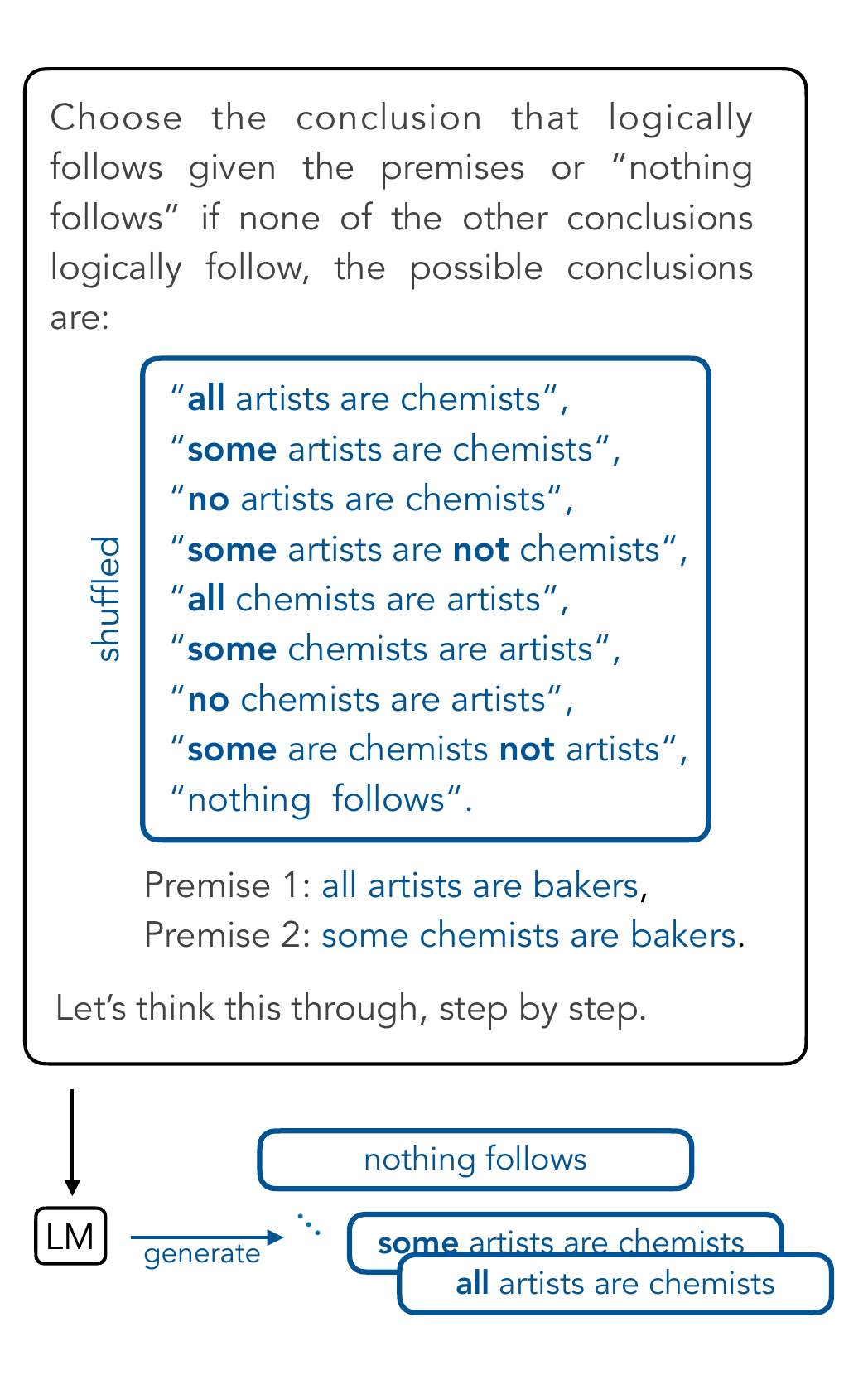
    }
    \caption{The zero-shot chain-of-thought prompt we use to assess LM
    syllogistic reasoning. The different parts of the prompt are grouped together
    for illustration purposes only; see also Figure~\ref{fig:prompts} in the
    Appendix for a purely textual representation of the prompt. }
    \label{fig:prompting}
\end{figure}
LMs trained with self-supervised objectives on large text corpora have been
instrumental in achieving high performance on a range of tasks. Some of the tasks in which LMs have shown promise have been referred to as reasoning tasks,
including commonsense reasoning, natural language inference, or question
answering (e.g., \citealt{chowdhery2022palm}). In this work, we focus more specifically
on deductive logical reasoning: drawing conclusions that \textit{must}, rather
than are likely to, be true given the premises, and where the inference is based
only on the premises, and does not rely on world knowledge. Unlike work on
datasets collected from textbooks or through crowdsourcing, we perform a well-controlled
analysis of a simple logical task for which there is a wealth of human data.

Several studies have benchmarked LMs on logical reasoning tasks \citep{Han2022-wn,Srivastava2022-dx,Wu2023-vm,Betz2020-fp,saparov2022language,Saparov2023-oi,Ye2023-zq}
and examined LM reasoning biases \citep{Dasgupta2022-ln,Razeghi2022-ix,Wu2023-lp,Thomas_McCoy2023-yz}.
\citet{saparov2022language} take a similarly controlled experimental approach to
ours (see also \citealt{Saparov2023-oi}), but they analyze LMs' performance on
formal logic rather than problems phrased in natural language as we do, and do not
compare their results to humans. The closest study to ours is \citet{Dasgupta2022-ln},
which demonstrates content effects in a number of logical reasoning domains, including
syllogisms. We extend their approach to study other aspects of syllogistic
reasoning.\looseness=-1

%% file: 4_method.tex
\section{Methods}
\begin{figure*}[t]
	\centering
	\includegraphics[width=\linewidth, clip, trim=11cm 7cm 9cm 8cm]{
		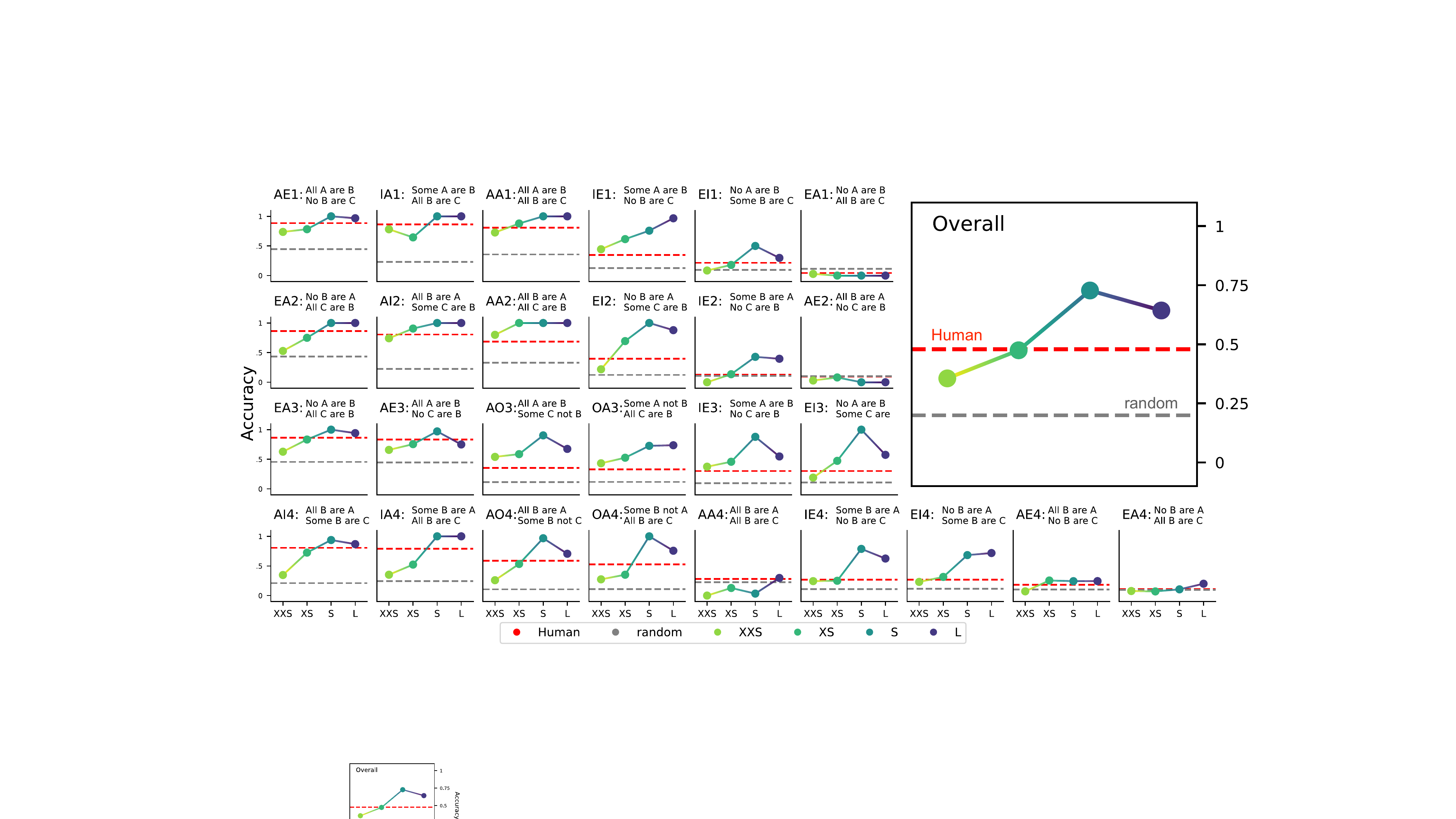
	}
	\caption{Accuracy of \mbox{PaLM~2} models, humans (red), and random guessing
	(grey). Random guessing accuracy differs by syllogism as some syllogisms have
	more than one valid conclusion. Syllogisms are partitioned into variable
	ordering (by row) and ordered by decreasing human accuracy from left to
	right. The top right inset shows the average accuracy across all syllogisms.
	Syllogisms are identified with the letters of the moods of the premises (Table~\ref{table:moods-variables},
	left) and the number associated with their variable ordering (Table~\ref{table:moods-variables},
	right).}
	\label{fig:hit-rate}
\end{figure*}

\subsection{Data}
The human behavioral data we use is drawn from \citet{Ragni2019-qo}, an online experiment where 139 participants
responded once to each of the syllogisms. In each trial, a participant was presented
with a syllogism and was instructed to choose among nine options: the eight
possible conclusions and ``nothing follows''. The experimental trials were preceded
by a brief training phase where participants were familiarized with the task.

Following \citet{Ragni2019-qo}, we generate syllogisms by replacing the abstract
terms (A, B, C) in each syllogism with one of 30 content triples chosen such that
there is no obvious semantic association between the terms (e.g., one of the
triplets included \emph{hunters}, \emph{analysts} and \emph{swimmers}; see Appendix~\ref{sec:content-words}
for the full list). This resulted in $64 \times 30 = 1920$ unique data points.

\subsection{Models and Inference}
\begin{figure}
	\centering
	\includegraphics[width=0.7\linewidth, clip, trim=25cm 11cm 22cm 12cm]{
		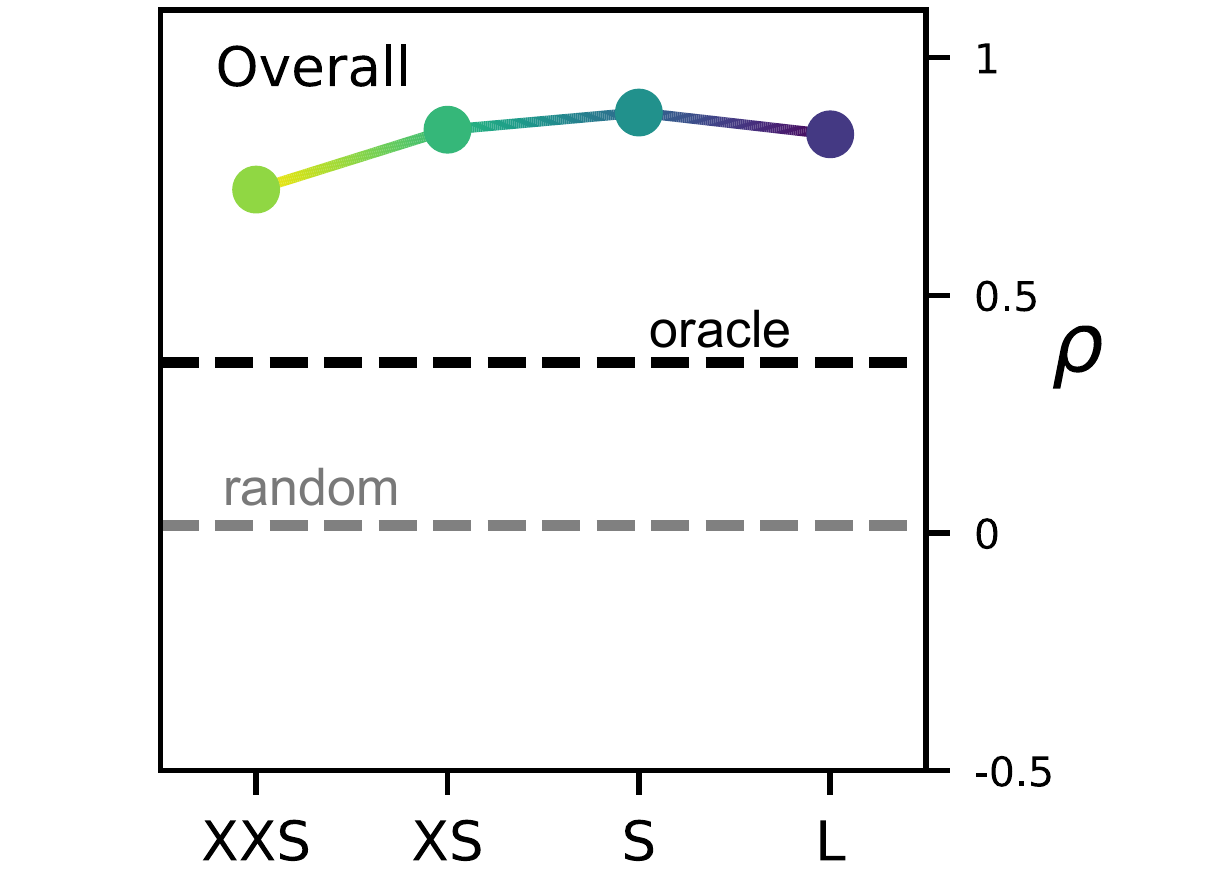
	}
	\caption{Correlation between \mbox{PaLM~2} models' predictions and human predictions.
	The oracle here is a logically correct reasoner that samples a response at
	random from all valid responses; the correlation of such an oracle with humans
	is relatively low as it does not mimic human errors.}
	\label{fig:human-corr}
\end{figure}
Most of our analyses focus on the \mbox{PaLM~2} family of LMs, which are
publicly available in four sizes (XXS, XS, S, and~L; ~\citealt{google2023palm2}).
These are transformer LMs trained on a
large corpus of multilingual web documents, books, code, mathematics and conversations. We also repeat all of our analyses for the 7B-, 13B- and 70B-parameter versions of the Llama~2
family of transformer models \cite{touvron2023llama}. Since, unlike for PaLM~2, we were unable to explore the different hyperparameters of our evaluation method for these models, we regard these results as preliminary and summarize them separately from the PaLM~2 results (Section~\ref{sec:llama} and Appendix~\ref{sec:llama-appendix}). All of the models we use
are pretrained only, without additional fine-tuning to match human preferences.
\begin{figure*}
	\centering
	\includegraphics[width=\linewidth, clip, trim=0cm 12cm 0cm 13cm]{
		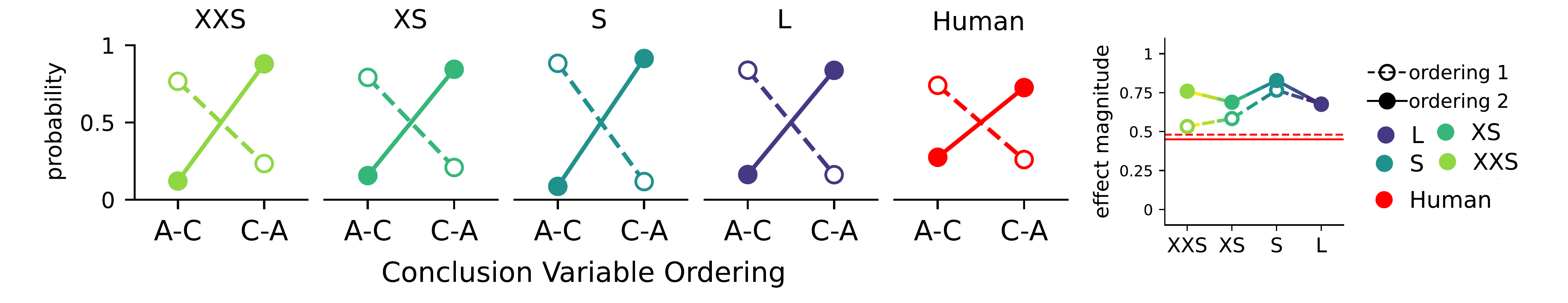
	}
	\caption{Variable ordering effects in PaLM~2 models and humans. \textbf{Left:} The marginal probabilities of A-C and C-A ordered
	conclusions.  \textbf{Right:} The
	magnitude of the variable ordering effect (the absolute value of the difference
	between the C-A probability and the A-C probability).}
	\label{fig:figural-effect}
\end{figure*}

Following the emerging standard practice for eliciting reasoning from LMs, we
use zero-shot ``chain-of-thought'' prompting, where the model is instructed to ``think
step by step'' \citep{kojima2022large,Wei2022-zv}. We speculate that the more explicit
reasoning process triggered by such prompts may more closely resemble the behavior
of human participants in experiments; for an analysis of alternative prompting
strategies that we explored before settling on this one, see Appendix~\ref{sec:cot-prompt-verbal}.
The prompt we use is illustrated in Figure~\ref{fig:prompting}. We randomize the
order of the conclusions in the prompt to control for LMs' sensitivity to answer
ordering \citep{Pezeshkpour2023-vk}.

For each of the 1920 reasoning problems, we estimated the distribution over
conclusions for each LM with a rejection sampling approach. Samples were rejected
if no conclusion was identified via uncased exact string match, and we took the
LMs' response to be the conclusion with the highest probability in this
distribution. Each distribution was estimated with 30 such proposals generated
with a temperature of 0.5 and a maximum decoding length of 75 tokens. See Appendix~\ref{sec:prompting}
for further details and an exploration of the impact of different prompts and decoding
parameters.

%% file: 5_results.tex
\section{Results\protect\footnote{An earlier version of this work, which was shared
on arXiv, reported slightly different results for the experiments discussed due
to a sampling bug.}}
\label{sec:results}

\subsection{Do LMs Reason Accurately?}
\label{sec:lms-logical}

We first examine the \mbox{PaLM~2} LMs' behavior on each of the 64 syllogism types
separately. In practice the LMs rarely produced the output ``nothing follows'', which is the
correct conclusion for 37 of the syllogisms. We return to this behavior briefly
in Section~\ref{sec:nothing-follows}, but in most of the following analyses, we
restrict ourselves to the 27 syllogisms that license conclusions other than ``nothing
follows'' (see Figure~\ref{fig:hit-rate} for the full pattern of results on each
of those 27 syllogisms). We compute the LMs' accuracy for each syllogism by
dividing the number of logically valid conclusions produced by the LM by the
total number of responses; note that some syllogisms have more than one valid conclusion
(as many as four) and so the random baseline in Figure~\ref{fig:hit-rate} varies by syllogism.

When averaged across all syllogisms, LM accuracy generally improves with scale, with
the two largest models exceeding human accuracy. The relationship between scale
and accuracy is not unambiguous, however: the largest model has somewhat lower
accuracy than the second-largest one. There is considerable variation across syllogisms;
for multiple syllogisms, accuracy is very low for all model sizes and can even decrease
as model size increases (this is the case, for example, for EA1: \emph{no A are B, all B are C}).

\subsection{Do LMs Reason Like Humans?}
\label{sec:lms-humanlike}Human accuracy averaged across all syllogisms is
roughly 50\% (Figure~\ref{fig:hit-rate}; red dashed line); as such, high LM
accuracy on this dataset does not necessarily imply humanlike reasoning. A comparison
by syllogism type reveals that the syllogisms that PaLM~2 models struggle with are
syllogisms that humans also find challenging, but the inverse is not true: multiple syllogisms
that are hard for humans are solved correctly by larger models. For example, for the
syllogism IE4 (\emph{some B are A, no B are C}), human accuracy is barely above chance,
but PaLM~2 Small and PaLM~2 Large are substantially more accurate.

\paragraph{Comparing the distribution over responses.}

So far we have focus on the proportion of correct responses. There are eight possible conclusions; is the distribution over all responses, including incorrect ones, similar across humans and PaLM~2 models? To
compute the probability distribution over conclusions for each syllogism, we
aggregate response counts for each syllogism and normalize them into a
probability distribution as in \citet{Khemlani_undated-xk}. We then correlate
the probability estimates from humans with the estimates from \mbox{PaLM~2}
models across the entire dataset (Figure~\ref{fig:human-corr}; for a by-syllogism
breakdown, see Figure~\ref{fig:human-corr-full} in Appendix~\ref{sec:human-corr}).
The correlation is fairly high across models, and is highest for \mbox{PaLM~2}
Small. 

\mbox{PaLM~2} Small and Large display both a high correlation with human responses
and a higher-than-human accuracy. This suggests that the miscalibration to human
data that models accrue due to higher accuracy is offset by a better fit to
humans elsewhere in the dataset. The next analyses test this hypothesis, zooming
in on two specific biases.

\paragraph{Variable ordering effects.}
\label{sec:figural-effects}
\begin{figure*}
    \centering
    \includegraphics[width=\linewidth, clip, trim=6cm 13cm 6cm 12cm]{
        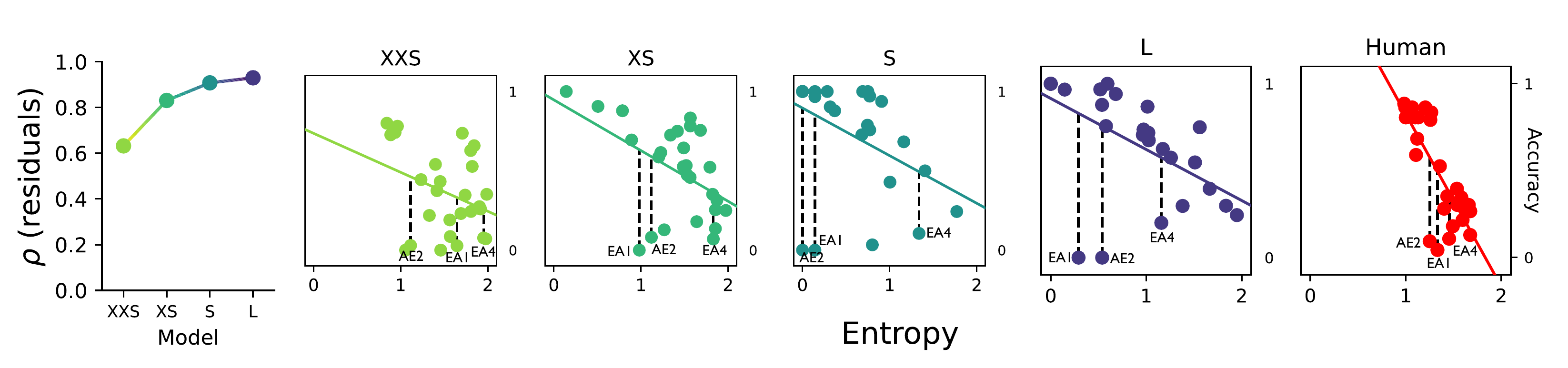
    }
    \caption{\textbf{Right:} Each syllogism plotted by accuracy (y-axis) and entropy (x-axis)
    and the regression line relating the two. Dashed lines black lines show the residuals
    for each of the top three human syllogistic fallacies. \textbf{Left:} The result of
    correlating \mbox{PaLM~2}'s regression residuals with residuals estimated from human
    data.}
    \label{fig:fallacies-examples}
\end{figure*}
Humans' syllogistic inferences are sensitive to variable ordering, even when the
ordering is logically irrelevant \citep{Johnson-Laird1978-jr}. Specifically,
humans produce more conclusions with an A-C variable ordering when reasoning in
response to a syllogism presented in ordering~1 (A-B, B-C); and they produce more conclusions with a C-A ordering when presented with a syllogism in
ordering~2 \mbox{(B-A, C-B)}.
We aggregate the human and LM responses across all \mbox{(A-B, B-C)} syllogisms
and across all \mbox{(B-A, C-B)} syllogisms separately and normalize the
aggregated response counts. All four PaLM~2 models show an ordering effect in the
same direction as humans (Figure~\ref{fig:figural-effect}, left). We compute the
magnitude of the effect as $|P\left (\text{A-C}
\right)-P\left(\text{C-A}\right)|$, where $P\left (\text{A-C}
\right)$ is the probability placed on conclusions with the order A-C. All models display a moderately larger bias
than humans (Figure~\ref{fig:figural-effect}, right). We do not find a clear trend
in the magnitude of the bias as model size increases; if anything, the largest model
shows a slightly weaker bias than the second-largest one.
\vspace{15pt}

\paragraph{Syllogistic fallacies.}
\label{sec:fallacies}

In general, humans are well-calibrated syllogistic reasoners---their accuracy is
inversely correlated with the entropy of their responses (Figure~\ref{fig:fallacies-examples}; see also \citealt{Khemlani2012-mj}). In other words, for most syllogisms where
humans give incorrect answers, the particular incorrect answers they give vary substantially
across individuals and trials. However, there are exceptions to this tendency:
in some cases, humans confidently and consistently choose a particular incorrect
answer (that is, low entropy coincides with low accuracy). For example, given
the syllogism \emph{no artists are bakers, all bakers are chemists}, humans
overwhelmingly respond with the logically invalid conclusion \emph{no artists
are chemists}; the correct conclusion, \emph{some chemists are not artists}, is produced
only 3\% of the time, and the distribution over responses elicited from humans for
this syllogism has one of the lowest entropies in the \citet{Ragni2019-qo}
dataset. We refer to such cases as \textit{syllogistic fallacies} \citep{Newsome2006-hh,Khemlani2017-zi}.

To identify potential fallacies in LMs, we fit a regression line relating
entropy (in nats) and accuracy, and then compute the distance from this line (the
residual) for each syllogism (Figure~\ref{fig:fallacies-examples}, right; for alternaitve
calibration measures, see \citealt{Guo2017-wm}). The top three human syllogistic
fallacies, defined as the top three outliers when plotting accuracy against
entropy, are also outliers for the PaLM~2 models. We also correlate the
residuals for all 27 syllogisms across humans and LMs, and find that larger models
display stronger correlations (Figure~\ref{fig:fallacies-examples}, left).

\paragraph{LMs avoid responding ``nothing follows''.}
\label{sec:nothing-follows}
\begin{figure}
    \centering
    \includegraphics[width=\linewidth, clip, trim=2.7cm 0cm 0cm 0cm]{
        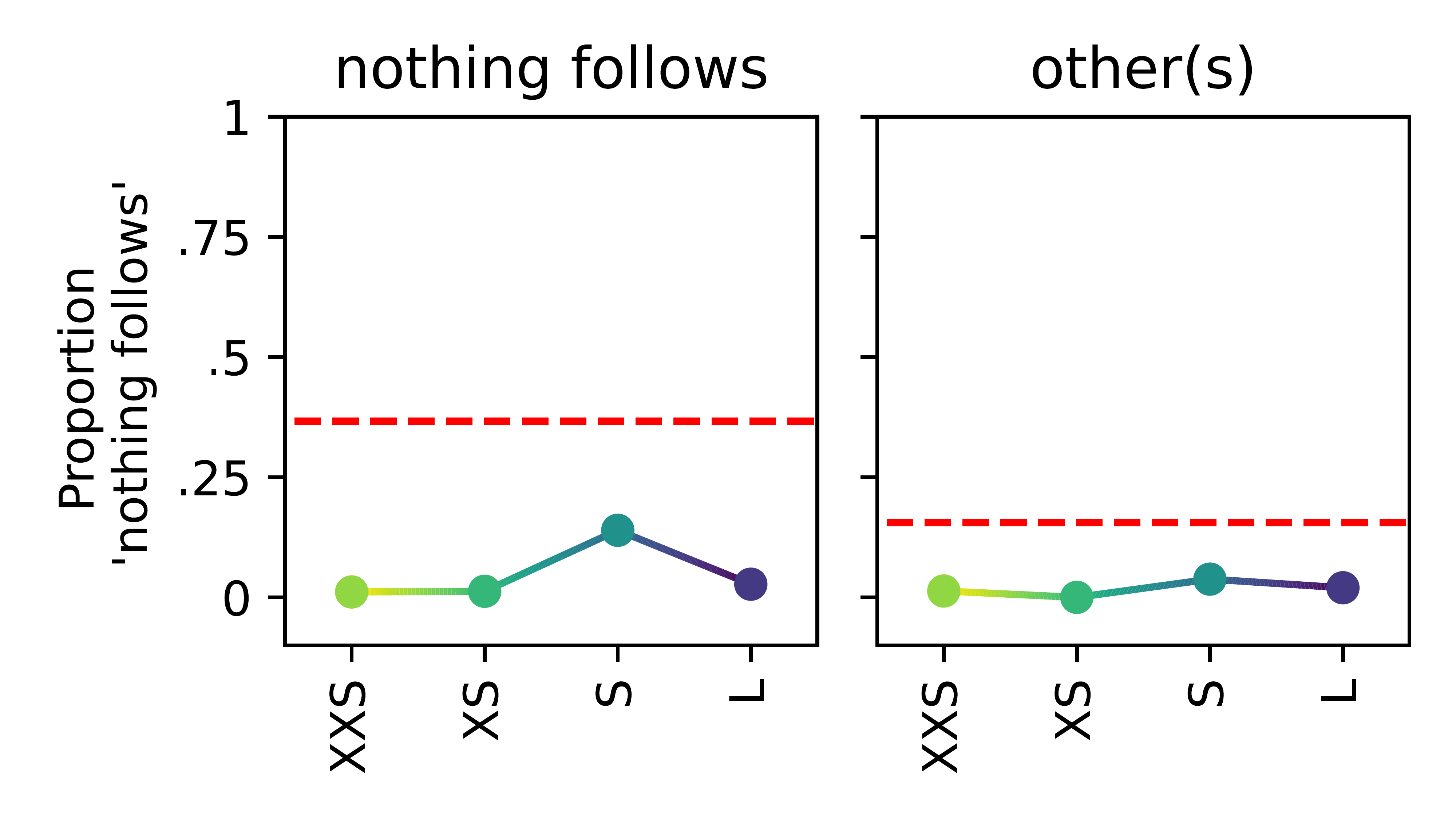
    }
    \caption{The proportion of ``nothing follows'' responses from humans and PaLM~2 models
    on the 37 syllogisms whose only valid conclusion is ``nothing follows'' (left)
    and the syllogisms that license conclusions other than ``nothing follows'' (right).}
    \label{fig:nothing-follows}
\end{figure}

An important divergence from human behavior is that LMs rarely produce the
response ``nothing follows'', even for the 37 syllogisms for which this is the correct
conclusion. Humans are also reluctant to conclude ``nothing follows''
\cite{Ragni2019-qo}, 
but the LMs' aversion to this response is much stronger than humans'---we
observe accuracies close to 0\% (Figure~\ref{fig:nothing-follows}). 
This issue is particularly severe with the zero-shot chain-of-throught prompting
method we use; in Appendix~\ref{sec:simplified-binary-evaluation}, we describe an
evaluation paradigm that can be used to elicit that conclusion, and leave
further analysis of this behavior to future work.

\subsection{Llama~2 Results}
\label{sec:llama}

Llama~2 models' overall accuracy, when aggregated across all syllogisms, was similar to human accuracy, with a
modest increase in accuracy as scale increases (Figure~\ref{fig:hit-rate_llama} in the Appendix). However, the breakdown by syllogism shows that this pattern masks substantial differences between humans and Llama~2: unlike PaLM~2 models, Llama~2 models struggle with some syllogisms that humans find easy, such as \emph{some A are B, all B are C}. Llama~2 models do, however, display a human-like variable ordering effect (Figure~\ref{fig:figural-effect_llama} in the Appendix). We refer the reader to Appendix~\ref{sec:llama-appendix} for plots of the results and additional analyses of Llama~2 models.

%% file: 6_mreasoner.tex
\section{Interpreting Language Models Using Mental Models Theory}
\label{sec:mentalmodels}
\begin{figure}
	\centering
	\includegraphics[width=0.82\linewidth, clip, trim=27cm 11cm 25cm 11cm]{
		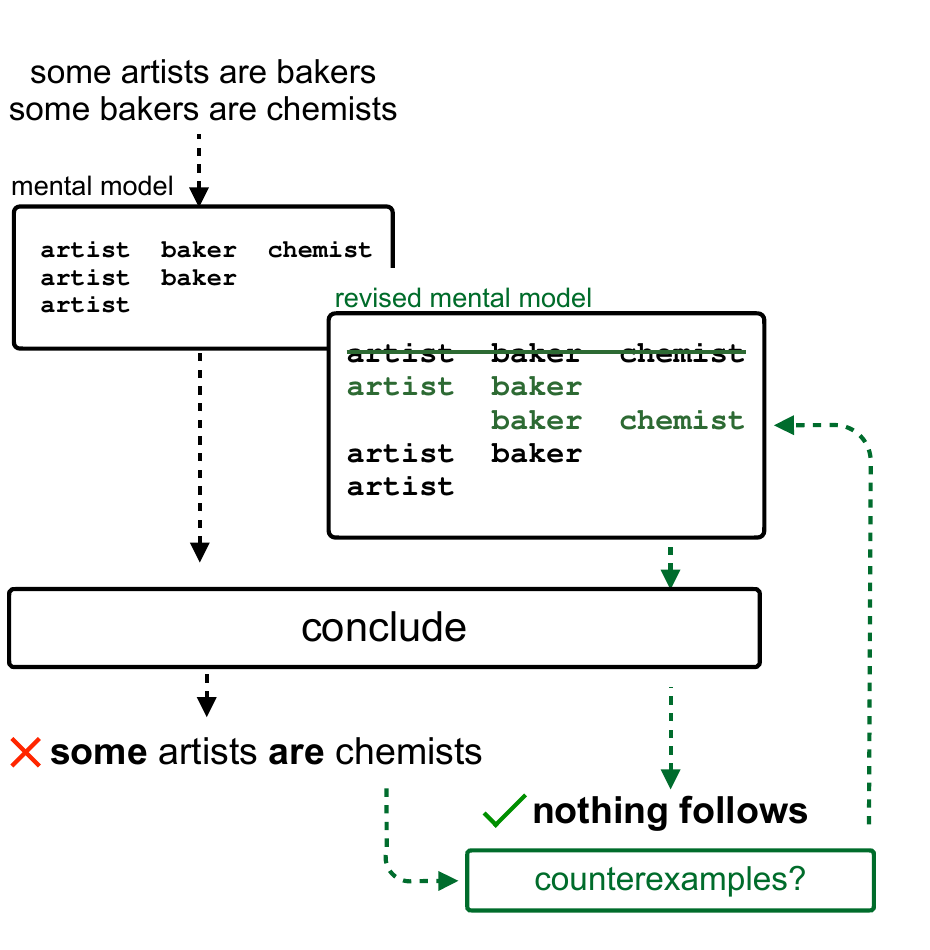
	}
	\caption{Schematic of mReasoner deducing an incorrect conclusion before
	finding counterexamples (``System~2'' processes shown in green) and updating to
	the correct conclusion, ``nothing follows''.}
	\label{fig:mreasoner-schem}
\end{figure}
We next analyze the behavior of PaLM~2 models using the Mental Models theory of human
logical reasoning \citep{Johnson-Laird1983-pl}, which has been developed over decades
to account for human data. The theory takes humans to be resource-limited and
simulation-based reasoners (\citealt{Craik1967-tx,Lake2017-hq,Lieder2019-dz,Johnson-Laird1983-pl},
i.a.), with a potentially high degree of variability across individuals. The implementation
we use---mReasoner\footnote{\url{https://github.com/skhemlani/mReasoner}} \citep{Khemlani2022-yw}---captures
these aspects of human reasoning with a small set of interpretable hyperparameters
that enable it to construct, refine, and draw conclusions from internal mental models
of the situations described in a syllogism.

Mental models consist of sets of entities instantiating the premises, where an
entity is represented by a conjunction of logical properties. For example,
Figure~\ref{fig:mreasoner-schem} illustrates a mental model corresponding
to the syllogism \emph{some artists are bakers, some bakers are chemists}. This
model consists of just three entities, the first of whom is an artist who is also
a baker and a chemist, the second is an artist and a baker who may or may not be
a chemist (this uncertainty is represented in the figure with a blank space), and
so on. The reasoner constructs and maintains its mental model with a set of
actions parameterized by four hyperparameters:
\begin{itemize}[leftmargin=*]
	\item \textbf{\texttt{LEN}} ($\lambda \in [1, \infty)$): The number of entities
		generated by the reasoner is sampled from a Poisson distribution with a mean
		of \textbf{\texttt{LEN}}.

	\item \textbf{\texttt{BROAD}} ($\epsilon \in [0, 1]$): Determines the set of
		individuals that mReasoner samples from. There are two possible sets: a broader
		set of all individuals consistent with the premises, and a smaller, canonical
		set of individuals consistent with the premises. The canonical sets were
		determined from human experiments (\citealt{Khemlani2022-yw}; for an example,
		see Figure~\ref{fig:mreasoner-cononical-sets} in Appendix~\ref{sec:mreasoner-app}). 

	\item \textbf{\texttt{SYSTM2}} ($\sigma \in [0, 1]$): The reasoner's propensity
		to reconsider its conclusion and search for counterexamples. Search is
		conducted by adding an entity to the model, moving a property from one entity
		to another, or decomposing one entity into two (these strategies are illustrated
		in Figure~\ref{fig:mreasoner-subroutines} in Appendix~\ref{sec:mreasoner-app}). 

	\item \textbf{\texttt{WEAKEN}} ($\omega \in [0, 1]$): Determines the model's
		reaction to finding a counterexample. The reasoner's options in this
		case are either to respond ``nothing follows'' or to weaken its response
		(i.e., amending erroneous \underline{global} conclusions such as \emph{all
		A are C} to weaker \underline{particular} conclusions such as \emph{some
		A are C}). When $\textbf{\texttt{WEAKEN}}$ is higher, mReasoner is more
		likely to weaken its response and less likely to answer ``nothing
		follows".
\end{itemize}
\begin{figure}
	\centering
	\includegraphics[width=0.5\linewidth, clip, trim=29cm 12.5cm 29cm 13.5cm]{
		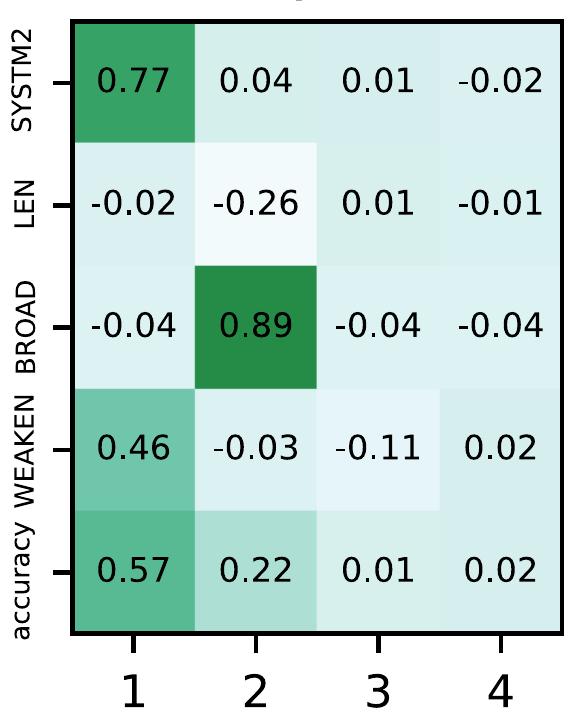
	}
	\caption{Correlations between the four principal components resulting from our analysis and mReasoner's original parameters (top four rows) as well as accuracy (bottom row).}
	\label{fig:mreasoner-conf}
\end{figure}

Figure~\ref{fig:mreasoner-schem} illustrates how mReasoner might process the syllogism
\emph{some artists are bakers, some bakers are chemists}. First, it constructs a
mental model, with length governed by \textbf{\texttt{LEN}} and content governed
by \textbf{\texttt{BROAD}}, consisting of the entities mentioned above: an artist-baker-chemist,
an artist-baker, and an artist. The conclusion \emph{some artists are chemists} is
consistent with this particular model (i.e., the first entity is both an artist
and a chemist). This conclusion is not true in every model that is consistent
with the premises, and as such it is not logically valid; however, if the reasoner
does not trigger a System 2 process, it will (incorrectly) take this conclusion
as valid and return it. Alternatively, with probability \textbf{\texttt{SYSTM2}}
mReasoner will scrutinize the conclusion by amending its model in an attempt to find
a counterexample. In this case, mReasoner successfully finds a counterexample by
breaking the first entity into two entities that are still consistent with the premises
but are not consistent with \emph{some artists are chemists}; consequently,
mReasoner corrects its answer to ``nothing follows''.

\paragraph{Mapping LM predictions onto cognitively meaningful dimensions.}
Syllogistic reasoning behavior is high-dimensional; in the set of syllogisms and
conclusions we consider, there are 27 syllogisms and eight possible responses to
each, for a total of 216 dimensions. We instantiate 1296 mReasoner models, one
for each point in a parameter grid, and analyze the 923 of them that finished simulations
before timing out (for the details of the parameter grid, see Table~\ref{table:mreasoner-grid} in Appendix~\ref{sec:mreasoner_grid}).
We evaluate each instance on each syllogism and represent the instance as a
vector in a 216-dimensional space. Finally, we use PCA to identify the top four
principal components in this space.

\paragraph{Characterizing the space of reasoning behaviors described by
mReasoner.}
Although mReasoner is characterized by four parameters, we find a single
principal component (PC~1) that captures~77\% of the variance in the model's behavior. This component loads heavily on $\textbf
{\texttt{SYSTM2}}$ and, to a lesser degree, on $\textbf{\texttt{WEAKEN}}$ (Figure~\ref{fig:mreasoner-conf}). Following the terminology of \citet{Khemlani_undated-xk}, we view this dimension as representing
\textit{deliberative} reasoning. Similarly, PC~2 loads heavily on
$\textbf{\texttt{BROAD}}$. This dimension, however, describes much less of the behavioral
variance of mReasoner.
\begin{figure}
	\centering
	\includegraphics[width=\linewidth, clip, trim=19cm 11cm 17cm 11cm]{
		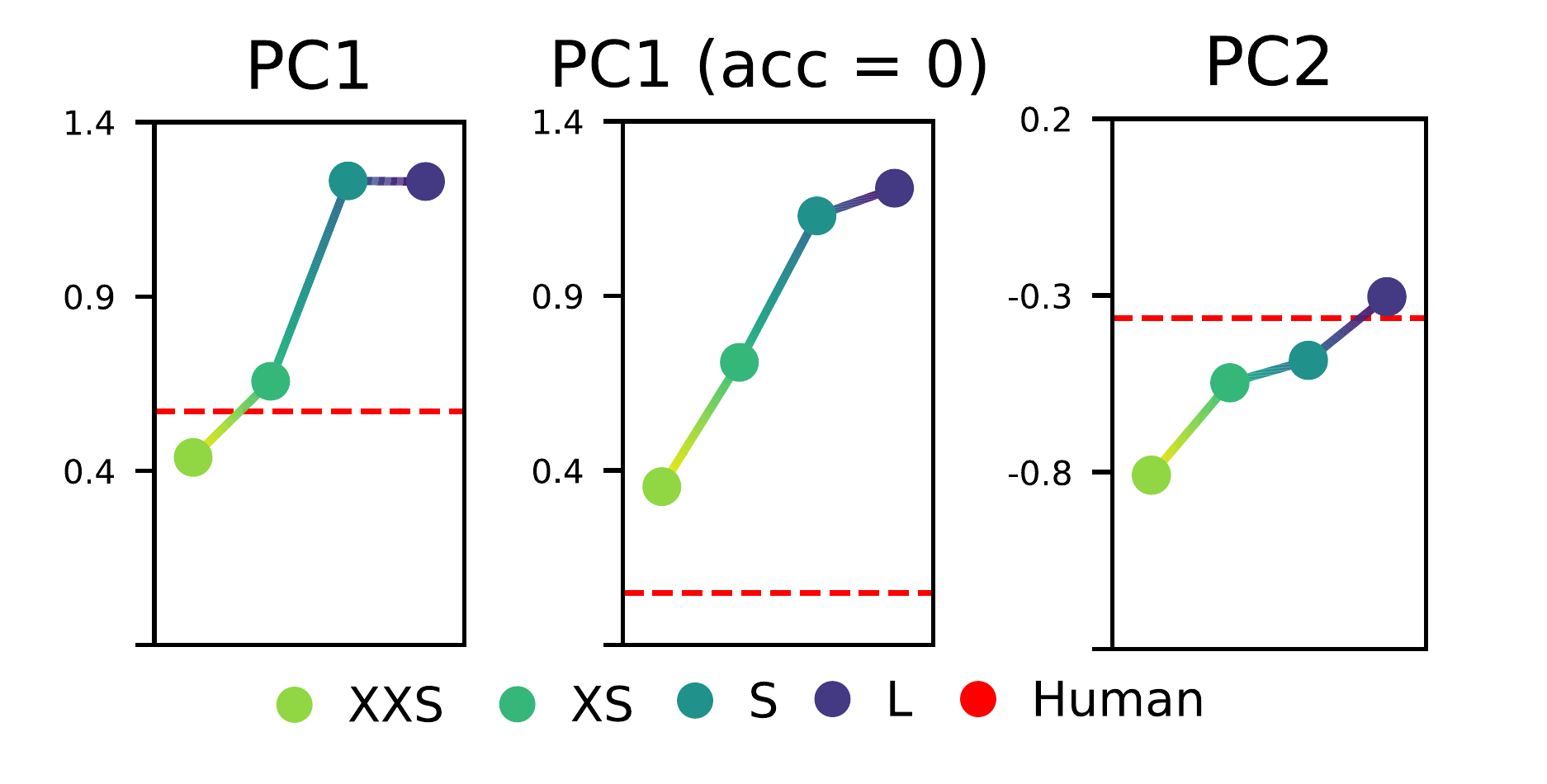
	}
	\caption{\textbf{Left and right:} Projecting PaLM~2 models onto the first two principal components of the feature space resulting from the behavior of simulations using mReasoner. \textbf{Center:} Projecting PaLM 2 models onto the same space when only their errors are taken into account.}
	\label{fig:mreasoner-pcs}
\end{figure}
\paragraph{LMs show signatures of deliberative reasoning.}
We project the 216-dimensional vectors describing the human data as well as the
behavior of each of our LMs into the PC space. This allows us to interpret the LMs'
behavior, in particular as model size increases, in terms of reasoning strategies
(Figure~\ref{fig:mreasoner-pcs}). We find that larger LMs behave more like
mReasoner instantiations with high $\textbf{\texttt{SYSTM2}}$ and
$\textbf{\texttt{WEAKEN}}$ values, as indicated by the fact that their first principal component is higher; in the terminology of \citet{Khemlani_undated-xk},
they show a stronger behavioral signature of deliberative reasoning.

\paragraph{Deliberative reasoning is partly dissociable from accuracy.}
PC~1 is strongly correlated not only with $\textbf{\texttt{SYSTM2}}$, but also with
accuracy. Can the changes in coordinates assigned to PaLM~2 be explained simply
by differences in accuracy? To test this, we repeat our analysis, this time setting
the probabilities of the correct answers to 0 for all mReasoner instantiations,
LMs and renormalizing (Figure~\ref{fig:mreasoner-pcs}, center). In
this control analysis, the accuracy of all models is 0\% (by design), but larger
models still display more deliberative reasoning. Here the deliberative
component (PC 1) has zero correlation with accuracy but a correlation of 0.6
with $\textbf{\texttt{SYSTM2}}$; correlations with all other parameters are below
0.15. This indicates that even the models' errors become more consistent with deliberative
reasoning.

%% file: 7_discuss.tex
\section{Discussion}

\paragraph{Human-like reasoning or accurate reasoning?}
Because of humans' systematic reasoning errors, syllogistic reasoning is a particularly
clear demonstration of the tension between the two central aims of artificial intelligence:
human-likeness and accuracy. We hypothesize that for most applications, accuracy
is more important than human-likeness; one notable exception is cognitive modeling,
where the goal is to better understand human reasoning by developing models that
reason like humans. We consider this application to be an important direction for
future work.

\paragraph{Why are LMs more accurate than humans?}
LMs learn from human-generated text, which is likely to reflect human beliefs and
biases; it is natural to hypothesize that the language modeling objective would
incentivize LMs to replicate those biases. We find only partial support for this
hypothesis. While the largest model's responses are indeed slightly more
correlated with human responses than the smaller ones, for some syllogisms where
humans reason very poorly, the models overcome human biases and reason
correctly. One possible explanation for this finding is that the data that PaLM 2
models were trained on includes not only natural language text, but also source
code \citep{chowdhery2022palm}, which may teach models to reason more
effectively. The effect of the composition of the LM's training corpus can be tested
in a controlled comparison in the future. %

\paragraph{Cognitive science for LM interpretation.}
We have used cognitive science to shed light on LM reasoning in two ways. First,
we used the biases documented in the cognitive psychology literature as hypotheses
for the biases that LMs might acquire. This approach is motivated by the
hypothesis that because LMs are trained on human-generated texts, which reflect human
biases and beliefs, they will be incentivized to replicate those biases to improve perplexity.
We found partial support for this hypothesis: larger LMs were more calibrated to human responses in some cases, in particular in our analysis of the correlation between accuracy and entropy (Section~\ref{sec:fallacies}).

The second and more novel way in which we use cognitive science is in interpreting LM
behavior using a computational cognitive model developed to explain human reasoning.
Under the assumption that LM reasoning follows the same heuristic strategies as
humans do (Section~\ref{sec:mentalmodels})---an assumption which, again, is informed by
the fact that LMs learn from text generated by humans---we concluded from this analysis
that LMs become more deliberative as their size increases.

\section{Conclusion}

Do LMs learn to reason correctly from self-supervised learning alone, even though
much of their training data was produced by humans, whose reasoning often
deviates from normative logic? We have addressed this question through a detailed
examination of the syllogistic reasoning behavior of the PaLM 2 family of LMs.
We find that the largest LMs make significantly fewer mistakes than humans
but still display systematic errors (Section~\ref{sec:lms-logical}), and that while
their mistakes are only partly aligned with human errors, LMs are susceptible to
several qualitative reasoning biases shown by humans (Section~\ref{sec:figural-effects}).

\section*{Acknowledgments}

We thank Andrew Lampinen for helpful discussion and Sangeet Khemlani for open-sourcing
MReasoner. TE is supported by the National Science Foundation Graduate Research
Fellowship under Grant No. 1745302.

\section*{Ethical Considerations and Limitations}
Part of this work's motivation is to extend the understanding of similarities
and differences between humans and current LMs, and we hope our work will have broader
positive impacts, such as facilitating cognitively informed and ethical model
development. Our results are limited, however, with challenges in directly comparing
LM and human behavior, and we comment on specific limitations below.

\paragraph{Eliciting LM reasoning.}
The space of possible ways to evaluate LMs on paradigms from human experiments is
fairly large. One can generate from the model \citep{aina-linzen-2021-language},
as we did; elicit meta-level judgements \citep{Hu2023-ru,Begus2023-do}; or simply
compare the probabilities assigned by the LM to possible continuations \citep{linzen2016assessing,Dasgupta2022-ln}.
Evaluations can be done in a zero-shot way, as we did, or in a few-shot way,
which may better approximate the training phase used in some human reasoning
experiments, such as \citet{Ragni2019-qo}; for discussion, see \citet{lampinen2022can}.
Finally, generative approaches can rely on a large set of possible prompts, and
can be used with or without ``chain-of-thought'' statements encouraging the
model to reveal its reasoning process \citep{kojima2022large}. Following
preliminary experiments (Appendix~\ref{sec:prompting}), we focused on zero-shot chain-of-thought;
a more systematic evaluation of the different elicitation approaches would be an
important direction for future work. 

\paragraph{The focus on Mental Models Theory.}
In Section~\ref{sec:mentalmodels}, we used a particular cognitive model, the Mental Models
Theory, to interpret LM reasoning behavior. This is not the only possible mechanism
that might underlie LM reasoning. Other accounts of human reasoning have argued
that people do, in fact, apply normative logic rules \citep{rips1994psychology},
perform probabilistic inference with constrained resources \citep{chater1999probability},
or combine probabilistic, heuristic and pragmatic reasoning \citep{Tessler2022-yi};
and it is possible that LMs reason in a way that does not match any of these theories.
We leave a systematic comparison of the fit of each of these theories to LM
reasoning for future work.

%% file: 8_appendix.tex
\section{Further Details on Syllogism Dataset}
\label{sec:content-words} Table~\ref{table:low-bias-content} displays the full
list of the content triples used in our experiments. The words in each triple were
chosen to have minimal semantic associations with each other.
\begin{table*}
	\centering
	\begin{tabular}{p{5cm}p{5cm}p{5cm}}
		\midrule actuaries, sculptors, writers & assistants, poets, scientists & athletes, assistants, chefs   \\
		chemists, drivers, dancers             & chemists, workers, painters   & clerks, butchers, athletes    \\
		dancers, bankers, riders               & doctors, riders, investors    & drivers, porters, chemists    \\
		farmers, surfers, writers              & gamblers, cleaners, models    & golfers, cyclists, assistants \\
		hunters, analysts, swimmers            & joggers, actors, carpenters   & linguists, cooks, models      \\
		linguists, skaters, singers            & managers, clerks, butchers    & miners, tellers, poets        \\
		models, tailors, florists              & nurses, scholars, buyers      & planners, sailors, engineers  \\
		riders, agents, waiters                & riders, novelists, linguists  & runners, opticians, clerks    \\
		scientists, novelists, florists        & skaters, barbers, cooks       & students, cashiers, doctors   \\
		students, hikers, designers            & surfers, painters, porters    & therapists, hikers, opticians \\
		\midrule
	\end{tabular}
	\caption{\label{table:low-bias-content}The 30 content word triples we use to
	construct syllogisms (e.g., for the first entry in the table, the variables A,
	B and C in the syllogism are replaced with \emph{actuaries}, \emph{sculptors}
	and \emph{writers}, respectively). The words in each triple were chosen to be
	minimally semantically associated with each other.}
\end{table*}

\begin{figure}
	\centering
	\includegraphics[width=.5\linewidth, clip, trim=0cm 0cm 0cm 0.5cm]{
		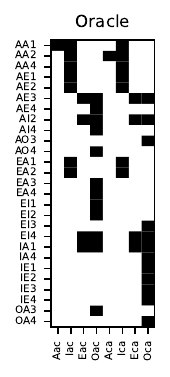
	}
	\caption{Valid conclusions for each syllogism. Conclusion identifiers show the
	conclusion mood (see Table~\ref{table:moods-variables}) followed by `ac' if the
	first variable in the conclusion is A and the second is C and `ca' in the opposite
	case.}
	\label{fig:valid-conclusions}
\end{figure}

\section{Prompting and Evaluation}
\label{sec:prompting} Before settling on the generative chain-of-thought
evaluation strategy that we focus on in this paper (described in detail in Section~\ref{sec:cot-prompt-verbal}),
we explored two additional strategies for eliciting and scoring syllogistic inferences
from LMs. First, we explored a multiple-choice approach, where, following the prompt,
we computed the mutual information between the prompt and each of the nine
possible conclusions (eight valid conclusions plus ``nothing follows''; Section~\ref{sec:multiple-choice});
and second, we explored a simplified binary discrimination approach, where, following
the prompt and a particular conclusion, we computed the mutual information
between the prompt and each of the strings ``valid'' and ``invalid'' (Section~\ref{sec:simplified-binary-evaluation}).
Of these three methods, chain-of-thought prompting achieved the highest accuracy
generally and had qualitatively similar performance across a range of
hyperparameters, so we use it in the main text. That being said, the binary discrimination
approach has the highest correlation with humans and is the only method that consistently
provides the response ``nothing follows'' when appropriate, and as such is a
promising method to explore in future work. The remainder of this appendix provides
additional details about the different elicitation methods and the variations on
those methods that we explored. All of the empirical results in this appendix
are based on PaLM~2.

\subsection{Generative Evaluation with a Zero-Shot Chain-of-Thought Prompt}
\label{sec:cot-prompt-verbal} The zero-shot chain-of-thought approach is illustrated
in Figure~\ref{fig:prompting}. We first describe the inference task to the model:
``Choose the conclusion that necessarily follows from the premises or ``nothing follows''
if none of the other conclusions logically \mbox{follow, ''}. We then define the
conclusion space, with the string ``the possible conclusions are: '' followed by
the list of all possible conclusions, including ``nothing follows''; the possible
conclusions are provided in a randomized order. Next, we provide the two premises
for the syllogism being queried in the format: ``Premise 1: $\texttt{PREMISE1}$,
\mbox{Premise 2: $\texttt{PREMISE2}$, ''}. Finally, we add the string ``Let's
think this through, step by step'', which is intended to instruct the LM to
produce a reasoning trace. We then generate from the LM, and determine for each
of the conclusions whether they appear in the text generated by the LM. The conclusion
that was detected most often, across content triples and samples, is taken to be
the answer produced by the model.

\subsubsection{Robustness to Prompt and Decoding Hyperparameters}
\label{sec:robustness}

The analyses presented in the main text are based on a decoding process in which
we sequentially generate 75 tokens from the LM, with a temperature of 0.5, and
take 30 such samples for each combination of syllogism type and content triple. Due
to compute limitations, we are unable to conduct a systematic exploration of
different variations on these hyperparameters for all model sizes; as such, we
focus on PaLM 2 XS. As in the main text, we only report accuracy for the 27
syllogisms that have valid conclusions, and exclude the syllogisms for which ``nothing
follows'' is the correct response.

\begin{figure*}
	\centering
	\includegraphics[width=\linewidth, clip, trim=10cm 5cm 6cm 4cm]{
		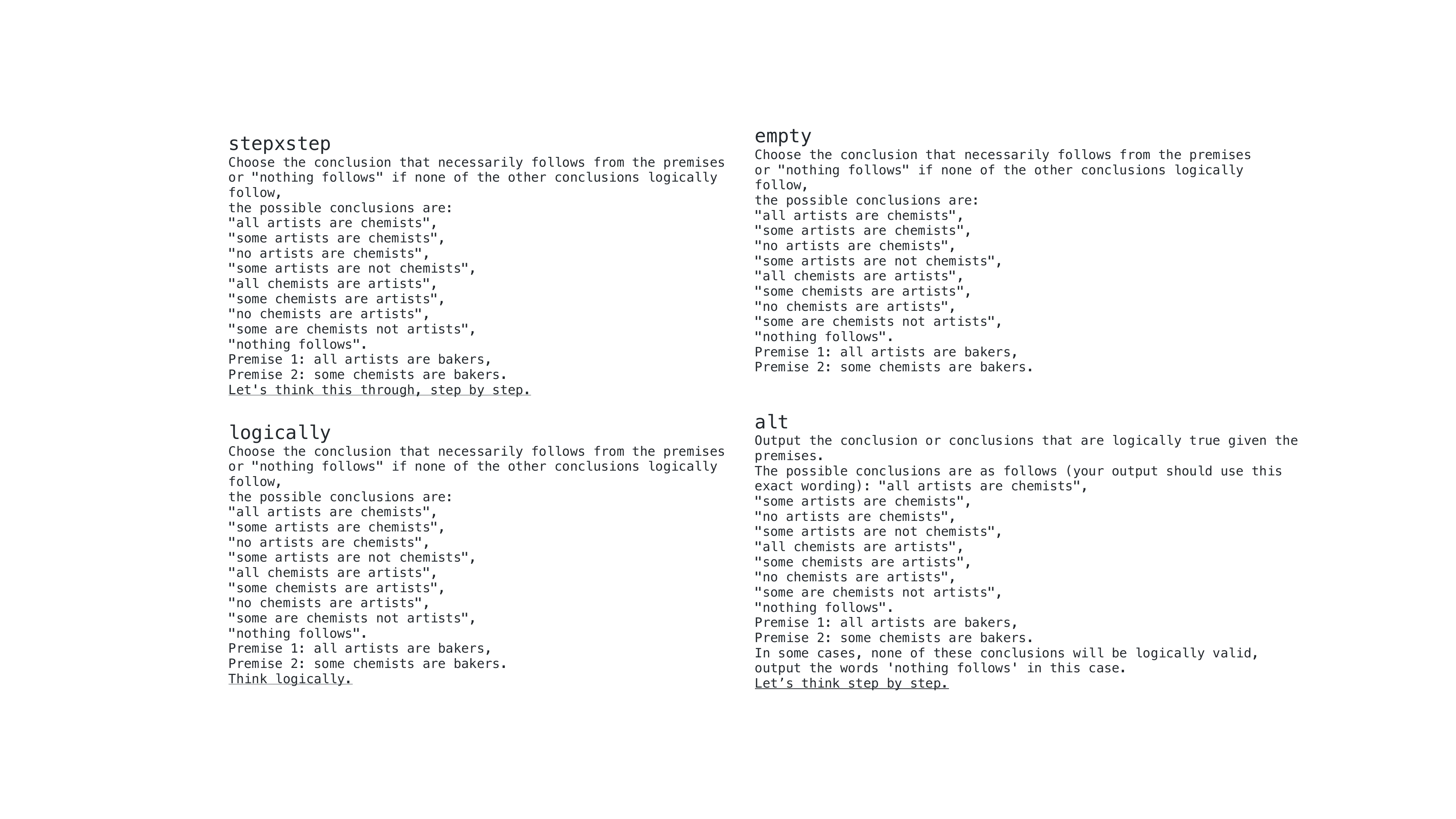
	}
	\caption{Variations on the prompt we used for the generative elicitation
	method; the prompt used in the main text is \texttt{stepxstep}.}
	\label{fig:prompts}
\end{figure*}

\paragraph{Prompts.}
In addition to the prompt we used in the main text, which we refer to as \texttt{stepxstep},
we consider three variations on this prompt (Figure~\ref{fig:prompts}):

\begin{enumerate}[leftmargin=*]
	\item \texttt{logically}: The same as \texttt{stepxstep}, except the zero-shot
		reasoning trigger ``Let's think this through, step by step'' is replaced
		by ``Think logically'' (like \texttt{stepxstep}, this prompt is inspired
		by a prompt from \citealt{kojima2022large}).

	\item \texttt{empty}: This prompt does not include any zero-shot reasoning trigger
		(that is, ``Let's think this through, step by step'' is replaced with
		the empty string).

	\item \texttt{alt}: We created this prompt in an attempt to mitigate the LMs'
		reluctace to produce ``nothing follows''; here the possibility of a ``nothing
		follows'' response is highlighted closer to the end of the prompt and in
		a more verbose way. This prompt also encourages the model to use the
		exact wording included in the prompt, and replaces ``Let's think this through,
		step by step'' with the slight variation ``Let's think step by step''.
\end{enumerate}

\begin{figure}
	\centering
	\includegraphics[width=1.1\linewidth, clip, trim=11cm 4cm 6cm 6cm]{
		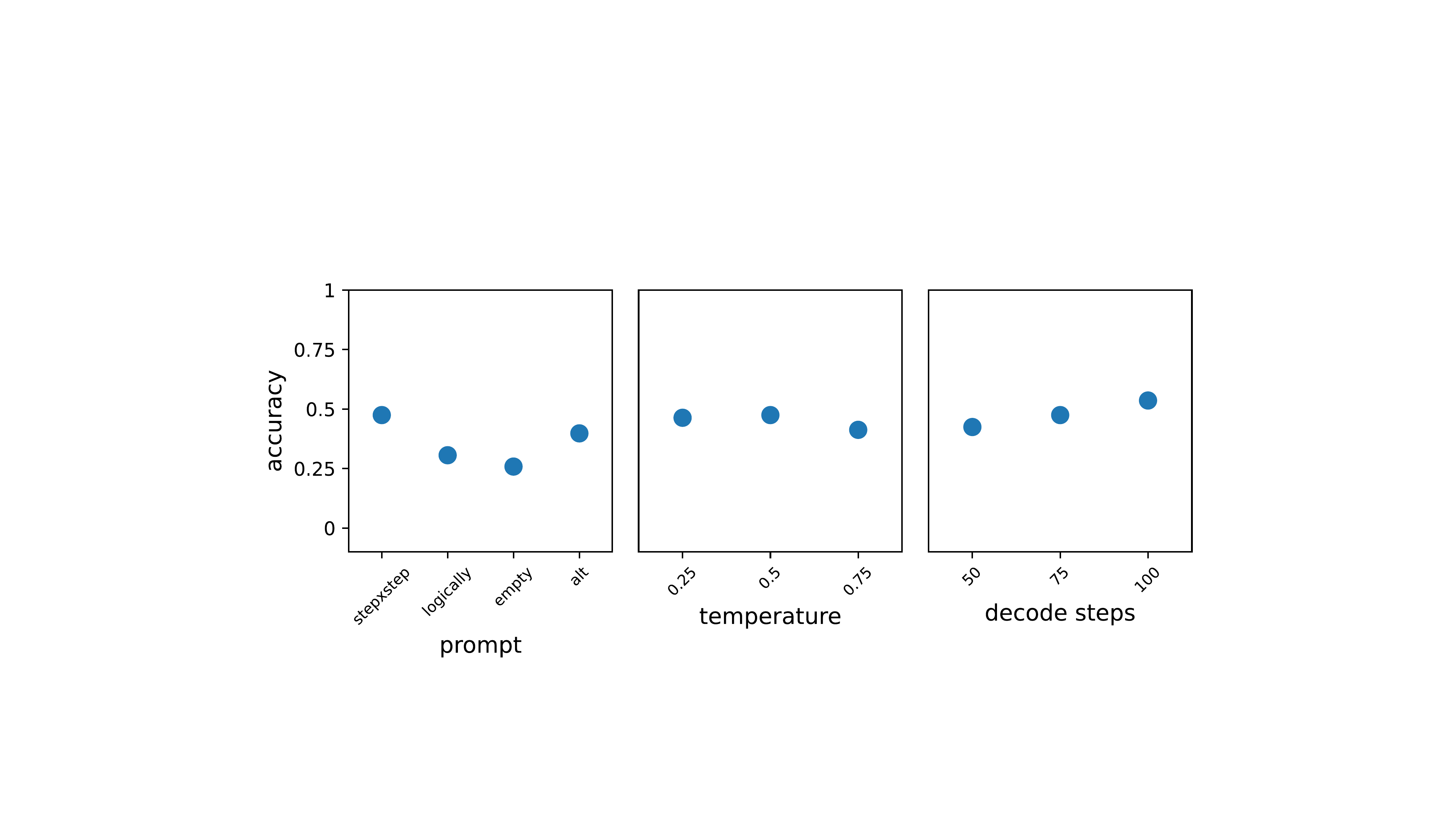
	}
	\caption{Accuracy for the chain-of-thought prompting method, with different prompts,
	temperatures and number of decoding steps.}
	\label{fig:hps}
\end{figure}

In the experiments varying the prompt, we hold the decoding temperature at 0.5
and the maximum number of decoded tokens at 75. We find that the prompt variants
show broadly similar patterns (Figure~\ref{fig:hps}), though \texttt{stepxstep}
achieves moderately higher accuracy than the other prompts.

\paragraph{Decoding hyperparameters}
Next, we hold the \texttt{stepxstep} prompt used in the main paper constant, and
independently vary decoding length and temperature. First, we use the
temperatures $\{0.25,0.5,0.75\}$, holding the decoding length at 75. Second, we
vary the number of tokens decoded between $50$, $75$ and $100$, keeping the
temperature at 0.5. Here, we observe a slight increase in accuracy as the number
of decoded tokens increases, which is expected (Figure~\ref{fig:hps}).

\subsection{Multiple-Choice Discriminative Evaluation}
\label{sec:multiple-choice}
\begin{figure*}
	\centering
	\includegraphics[width=\linewidth, clip, trim=12cm 4cm 12cm 3cm]{
		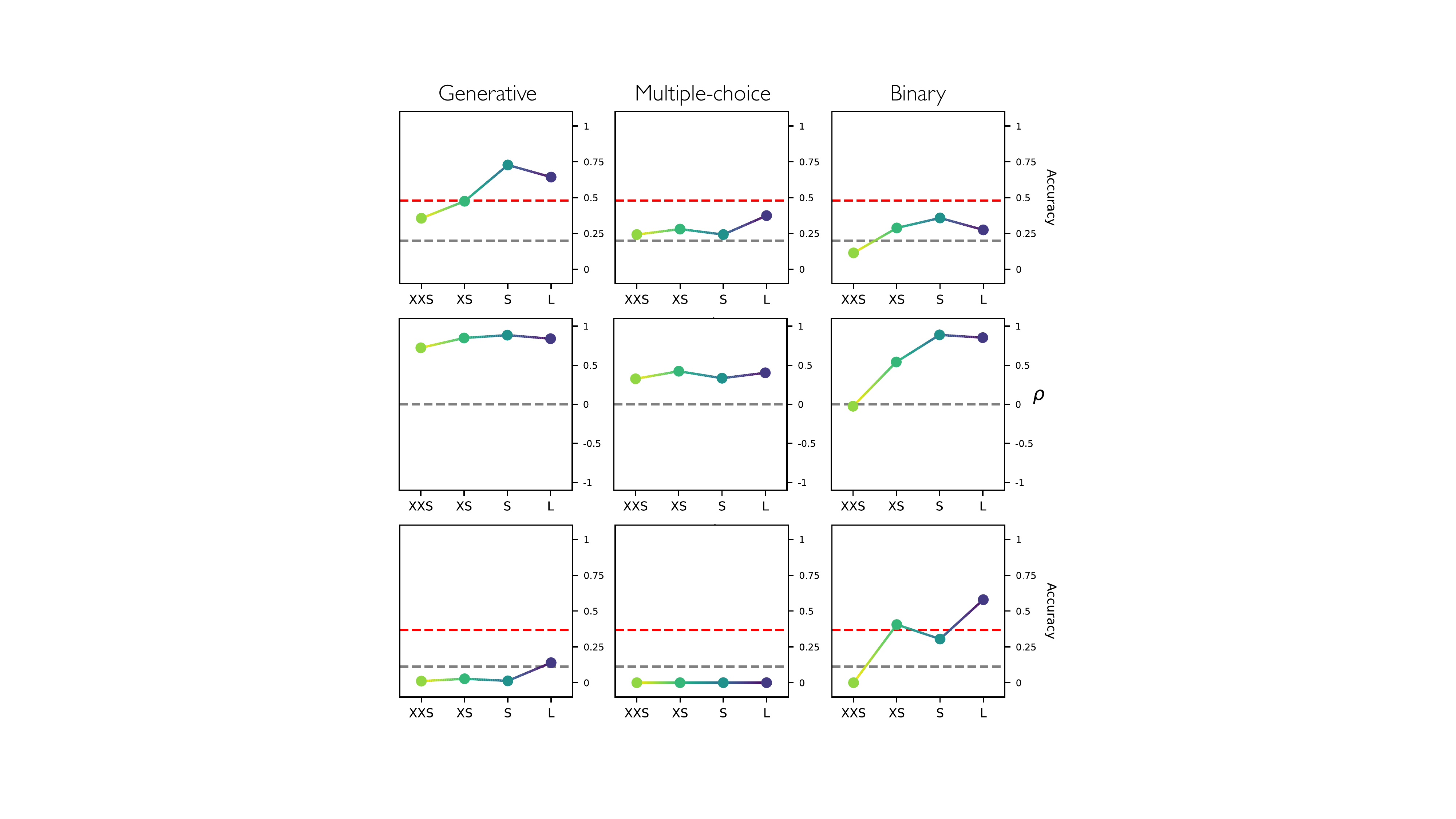
	}
	\caption{Comparison across reasoning elicitation methods with PaLM~2 models: the CoT generation
	method used in the main text (generative), as well as the binary and multiple-choice
	methods. We show accuracy among syllogisms with a valid
	conclusion (top), correlation with humans (middle), and accuracy among
	syllogisms where the correct response is ``nothing follows'' (bottom). The accuracy of
	the generative method is highest on the valid syllogisms, but the
	binary discrimination method achieves markedly higher accuracy on
	the ``nothing follows'' syllogisms. Both outperform the multiple-choice method
	substantially.}
	\label{fig:prompting-strategies}
\end{figure*}
In this approach to evaluating LM reasoning, we replace the generative evaluation
with a discriminative scoring of each of the possible conclusions. The prompt is
very similar: we remove the zero-shot chain-of-thought trigger from \texttt{stepxstep}
and replace it with ``The conclusion that necessarily follows is: '', then feed the
prompt to the models and score each of the conclusions. To normalize for the
idiosyncratic features of each conclusion, such as its length and prior probability,
we use the \textit{mutual information} between the prompt (p) and the conclusion
(c) as the score~\citep{Holtzman2021-vq}:
\begin{align}
	\text{MI}(\text{c};\text{p}) = \log P(\text{c}|\text{p}) - \log P(\text{c}|\text{``''})
\end{align}
We then renormalise these scores to compute a distribution over the conclusions (indexed
by $i$):
\begin{equation}
	P\left(\text{c}_{i}\right)=\frac{\exp \left(\mathrm{MI}\left(\text{c}_{i}; \text{p}\right)\right)}{\sum_{j}\exp
	\left(\mathrm{MI}\left(\text{c}_{j}; \text{p}\right)\right)},
\end{equation}
and take the conclusion with the highest $P\left(\text{conclusion}_{i}\right)$
to be the LM's prediction for a given combination of syllogism and content
triple. Results obtained using this method are shown in Figure~\ref{fig:prompting-strategies}.

\subsection{Simplified Binary Evaluation}
\label{sec:simplified-binary-evaluation} While the multiple-choice format is most
similar to the paradigm used in human experiments, it poses a significantly
harder task than simple binary discrimination \citep{Dasgupta2022-ln}, which may
be more sensitive. In the validity discrimination evaluation method, we present
the LM with the prompt ``Is this conclusion valid given the premises:'' followed
by the premises and a single conclusion (we refer to the concatenation of the
prompt and $\text{conclusion}_{i}$ as $\text{prompt}_{i}$ below). We do this for
all eight possible conclusions (omitting ``nothing follows''). We, again, use the
mutual information to score and compute the binary probability of ``valid'' as:

\begin{align}
	 & P\left(\text{``valid''}|\text{c}_{i}\right)=                                                                                                                                                                                                  & \hspace{5cm}\nonumber \\
	 & \frac{\exp \left(\mathrm{MI}\left(\text{``valid''};\text{p}_{i}\right)\right)}{ \exp \left(\mathrm{MI}\left(\text{``valid''};\text{p}_{i}\right)\right) + \exp \left(\mathrm{MI}\left(\text{``invalid''};\text{p}_{i}\right)\right)}\nonumber &
\end{align}

We compute discrete conclusion decisions by normalizing $P(\text{``valid''})$
for each conclusion into a probability distribution:
\begin{equation}
	P\left(\text{c}_{i}\right)=\frac{P\left(\text{``valid''}|\text{c}_{i}\right)}{\sum_{j}P\left(\text{``valid''}|\text{c}_{j}\right)}
	, \label{eg:valid_discrimination_renormalized}
\end{equation}

and taking the conclusion with the largest probability according to Equation~\ref{eg:valid_discrimination_renormalized}
to be the LM's selected conclusion for a syllogism (the conclusion most likely to
be valid according to the LM). In this approach, the LM's prediction is taken to
be ``nothing follows'' if $P\left(\text{``valid''}|\text{conclusion}\right)$
does not exceed 50\% for any of the conclusions. We note that this method is the
only one that successfully elicits ``nothing follows'' conclusions for a
substantial proportion of the syllogisms (Figure~\ref{fig:prompting-strategies}).

\section{By-Syllogism Correlations with Human Responses}
\label{sec:human-corr} 
\begin{figure*}
	\centering
	\includegraphics[width=\linewidth, clip, trim=10cm 8cm 9cm 6cm]{
		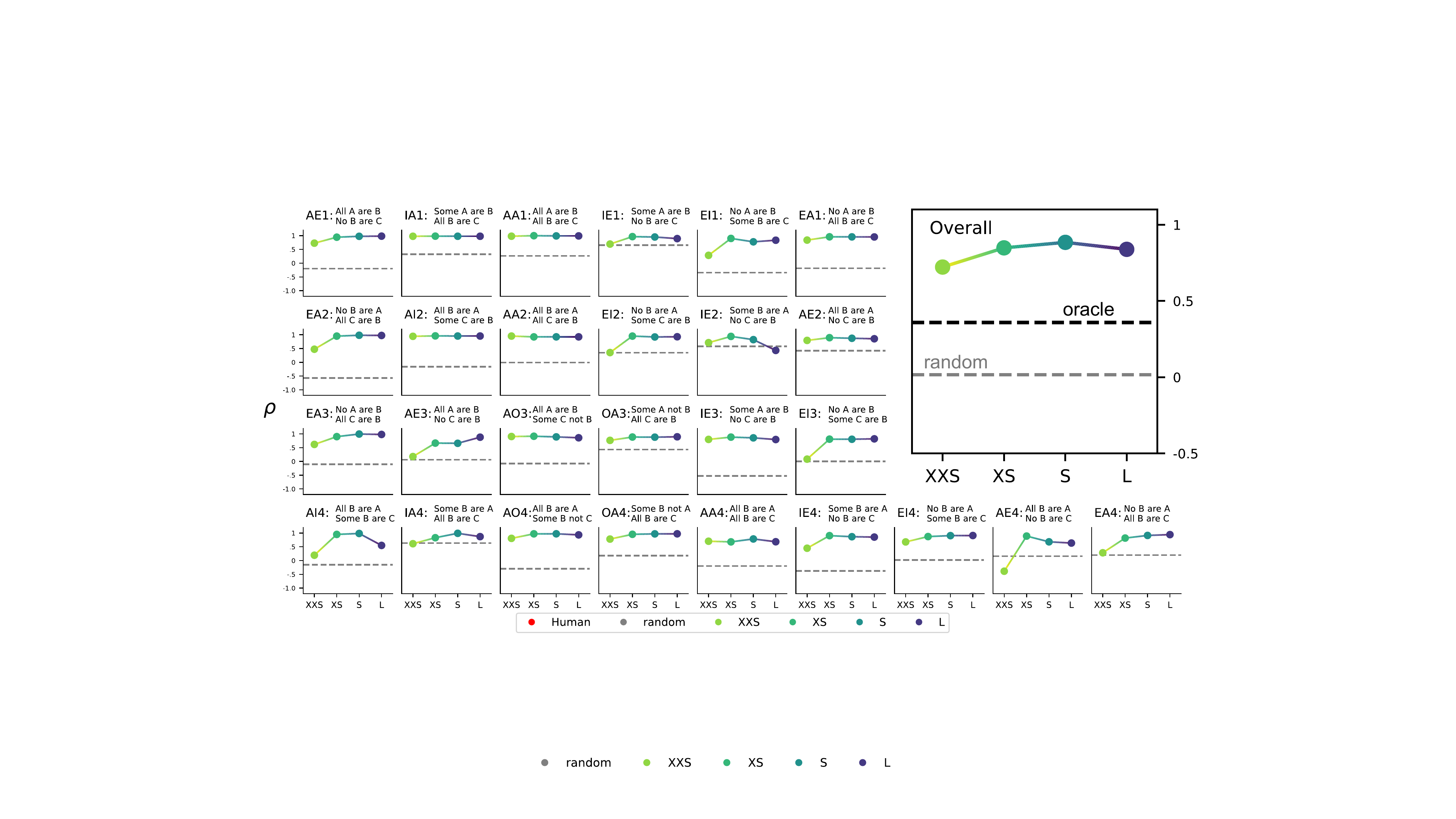
	}
	\caption{Correlation between PaLM~2 models' distribution over responses and the probabilities
	derived from normalizing human responses, broken down by syllogism. Syllogisms
	are partitioned into variable ordering type (by row) and ordered by decreasing
	human accuracy from left to right. Chance performance (dashed grey line)
	reflects random guessing. The top right inset shows correlation across the entire
	dataset.}
	\label{fig:human-corr-full}
\end{figure*}

Figures~\ref{fig:human-corr-full} provides correlations
between LMs and humans at the individual syllogism level. While larger LMs are generally
more human-like, we observe a diversity of relationships between model scale and
human-likeness, including cases such as IE2 where larger models are in fact less
correlated with humans.

\section{Mental Models Simulations: Additional Details}
\label{sec:mreasoner-app}
\subsection{Model details}
\begin{figure*}
	\centering
	\includegraphics[width=\linewidth, clip, trim=7cm 13cm 6cm 13cm]{
		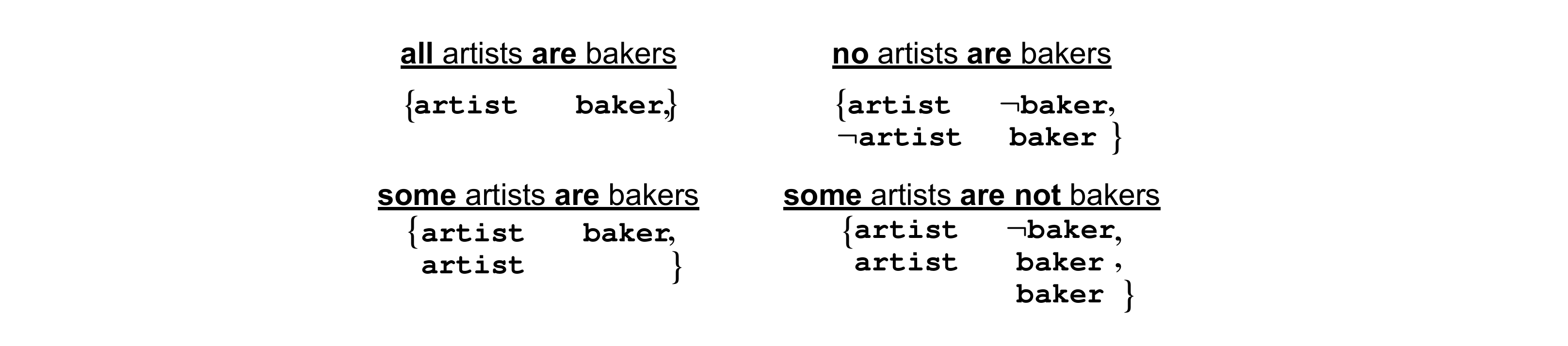
	}
	\caption{The ``canonical sets'' used by mReasoner. The canonical set for a syllogism
	depends on the moods of the syllogism's premises. We show the possible
	individuals each premise contributes to a syllogism's canonical set here for
	the hypothetical content words \emph{artists} and \emph{bakers}.}
	\label{fig:mreasoner-cononical-sets}
\end{figure*}

\begin{figure*}
	\centering
	\includegraphics[width=\linewidth, clip, trim=7cm 13cm 6cm 13cm]{
		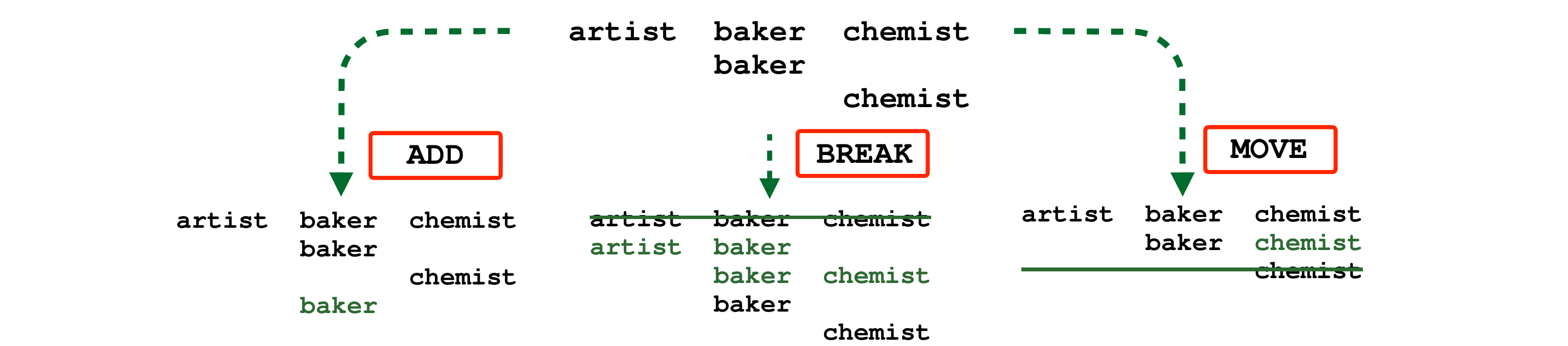
	}
	\caption{Subroutines used by mReasoner to revise mental models in order to
	check for counterexamples. We denote these subroutines as \textbf{\texttt{ADD}},
	\textbf{\texttt{BREAK}}, and \textbf{\texttt{MOVE}}, following \cite{Khemlani2022-yw}.
	\textbf{\texttt{ADD}} adds one more entity to a mental model. \textbf{\texttt{BREAK}}
	decomposes an entity's properties into constituent entities with subsets of those
	properties. \textbf{\texttt{MOVE}} simply moves a property from one entity to
	another.}
	\label{fig:mreasoner-subroutines}
\end{figure*}

This section provides additional details on mReasoner. Figure~\ref{fig:mreasoner-cononical-sets}
shows an example of the ``canonical sets'' that mReasoner uses to heuristically sample
entities, and Figure~\ref{fig:mreasoner-subroutines} illustrates the subroutines
used to revise mental models.

\subsection{mReasoner instantiations}
\label{sec:mreasoner_grid} We instantiate one mReasoner model for every
parameter vector in the grid shown in Table~\ref{table:mreasoner-grid}. This resulted
in a total of 1,296 models. As the model's reasoning process is stochastic, we evaluate
each model 100 times for each syllogism to estimate the distribution over responses.
Due to resource constraints, we discarded models that did not finish these 100 iterations
in 60 seconds, leaving us with 923 models spaced relatively evenly over the grid
(i.e., this timeout criterion did not systemtically favor some hyperparameter values).

Each of the 923 models is represented by a 216-dimension vector, with eight possible
conclusions for each of the 27 valid syllogisms ($27 \times 8 = 216$). We
perform a standard PCA---the probabilistic PCA of \citet{Tipping1999-og} on the centered
dataset, using scikit-learn \citep{scikit-learn}---on these 923 vectors.

\begin{table}
	\centering
	\begin{tabular}{r|cccccc}
		\textbf{\texttt{LEN}}    & 2.0 & 2.5 & 3.0 & 3.5 & 4.0 & 4.5 \\
		\textbf{\texttt{BROAD}}  & 0.0 & 0.2 & 0.4 & 0.6 & 0.8 & 0.9 \\
		\textbf{\texttt{SYSTM2}} & 0.0 & 0.2 & 0.4 & 0.6 & 0.8 & 0.9 \\
		\textbf{\texttt{WEAKEN}} & 0.0 & 0.2 & 0.4 & 0.6 & 0.8 & 0.9 \\
	\end{tabular}
	\caption{Parameter grid used to instantiate our mReasoner models.\label{table:mreasoner-grid}}
\end{table}

\section{Llama~2: Additional Results and Plots}
\label{sec:llama-appendix} 
\begin{figure*}
	\centering
	\includegraphics[width=\linewidth, clip, trim=4cm 13cm 3cm 12cm]{
		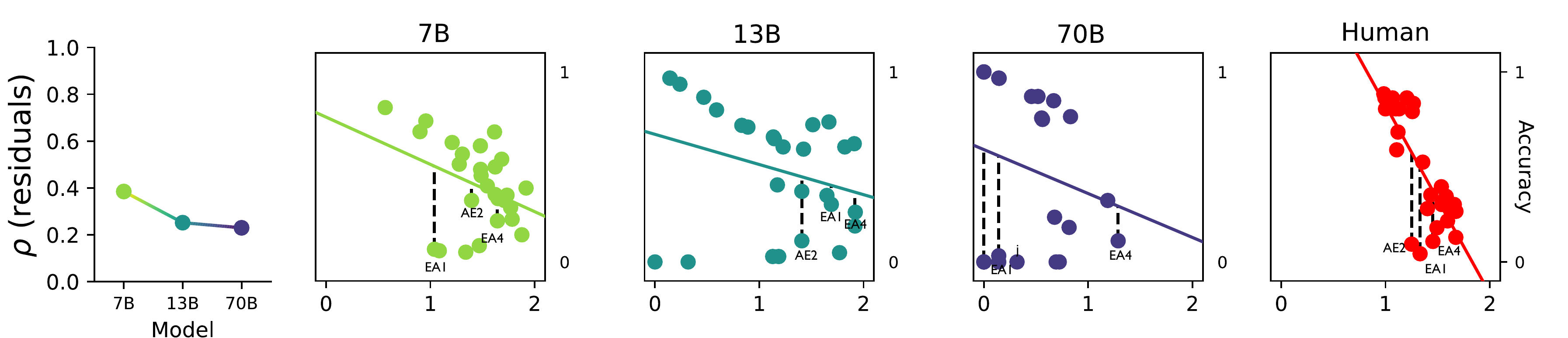
	}
	\caption{Analysis of Llama 2 models' handling of syllogistic fallacies. \textbf{Right:} Each syllogism plotted by accuracy (y-axis) and entropy (x-axis)
	and the regression line relating the two. Dashed lines black lines show the residuals
	for each of the top three human syllogistic fallacies. \textbf{Left:} The result of
	correlating Llama 2's residuals with residuals estimated from human data.}
	\label{fig:fallacies-examples_llama}
\end{figure*}

\begin{figure*}
	\centering
	\includegraphics[width=\linewidth, clip, trim=10cm 8cm 9cm 6cm]{
		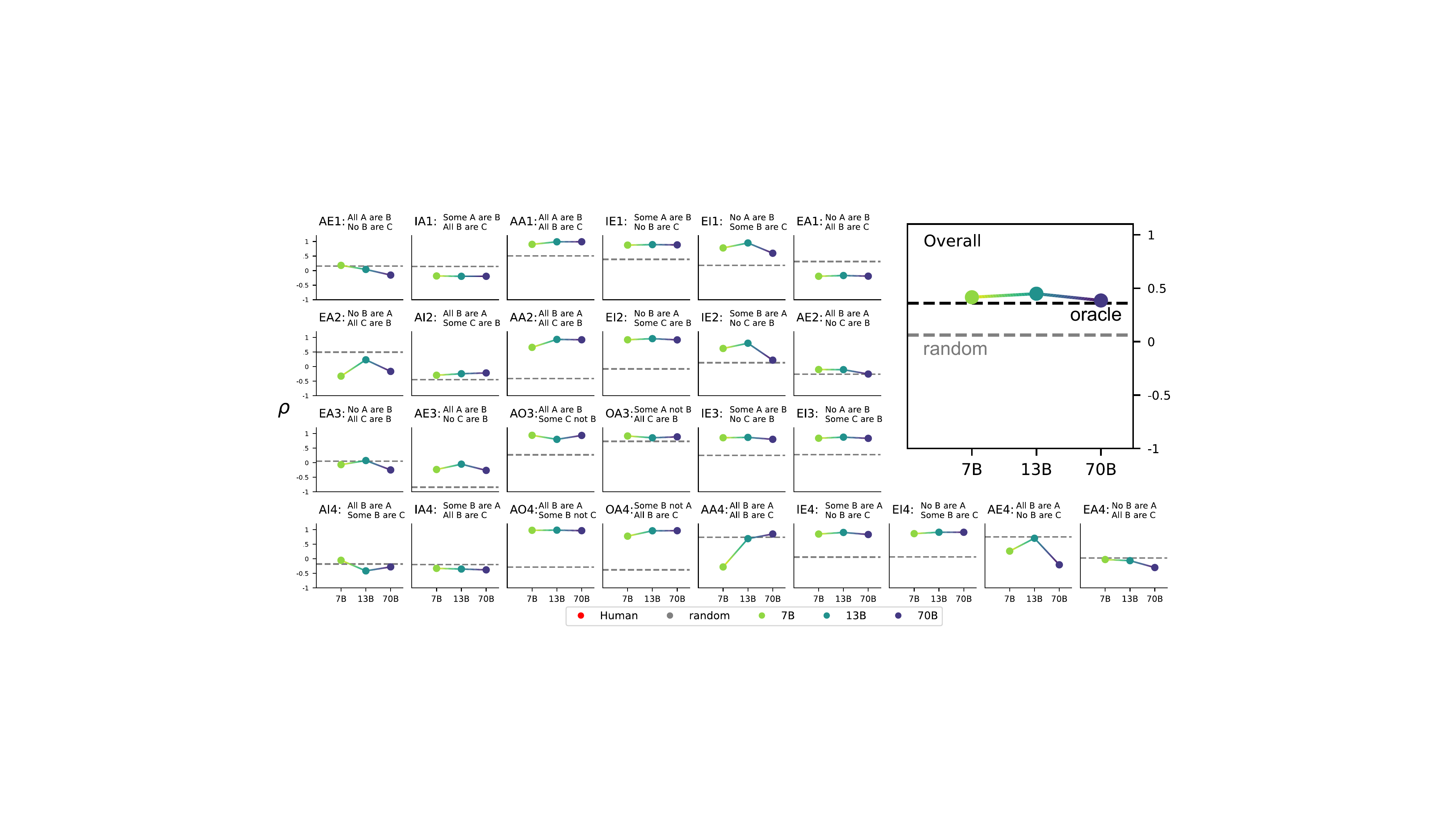
	}
	\caption{Correlation between the Llama~2 model's distribution over responses and
	the probabilities derived from normalizing human responses, broken down by syllogism.
	Syllogisms are partitioned into variable ordering type (by row) and ordered
	by decreasing human accuracy from left to right. Chance performance (dashed
	grey line) reflects random guessing. The top right inset shows correlation
	across the entire dataset.}
	\label{fig:human-corr-full_llama}
\end{figure*}

As we mentioned in Section~\ref{sec:llama}, while the overall accuracy of Llama~2 models is similar to that of humans, this aggregate pattern masks large discrepancies with humans, and in particular poor accuracy on some syllogisms where humans rarely make mistakes, as well as high accuracy on syllogisms that humans struggle with. Consequently, these models exhibit a substantially lower correlation
with human behavior across the board (Figure~\ref{fig:human-corr-full_llama}) than do PaLM~2 models (cf. Figure~\ref{fig:human-corr-full}). Llama~2 models also demonstrate a slight decrease in correlation with human
behavior as a function of model size in our analysis of the correlation between accuracy and entropy in syllogistic fallacies (Figure~\ref{fig:fallacies-examples_llama}; cf. Figure~\ref{fig:fallacies-examples} for PaLM~2 models).

\begin{figure*}
    \begin{subfigure}[b]{0.5\textwidth}
	\centering
	\includegraphics[width=\textwidth, clip, trim=2.7cm 0cm 0cm 0cm]{
		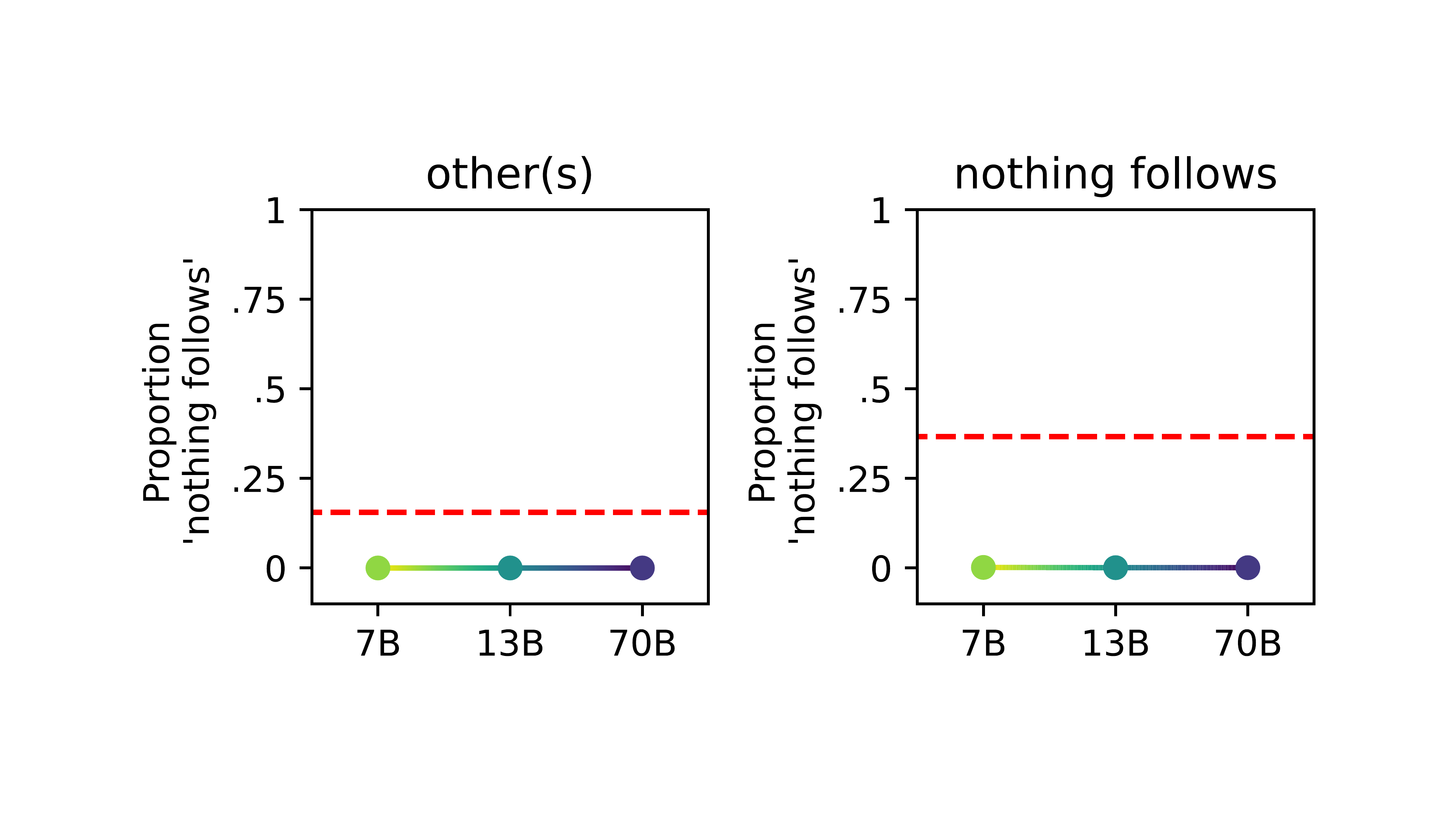
	}
	\caption{\label{fig:nothing-follows_llama}}
	
\end{subfigure}\begin{subfigure}[b]{0.5\textwidth}
	\centering
	\includegraphics[width=\textwidth, clip, trim=19cm 11cm 17cm 11cm]{
		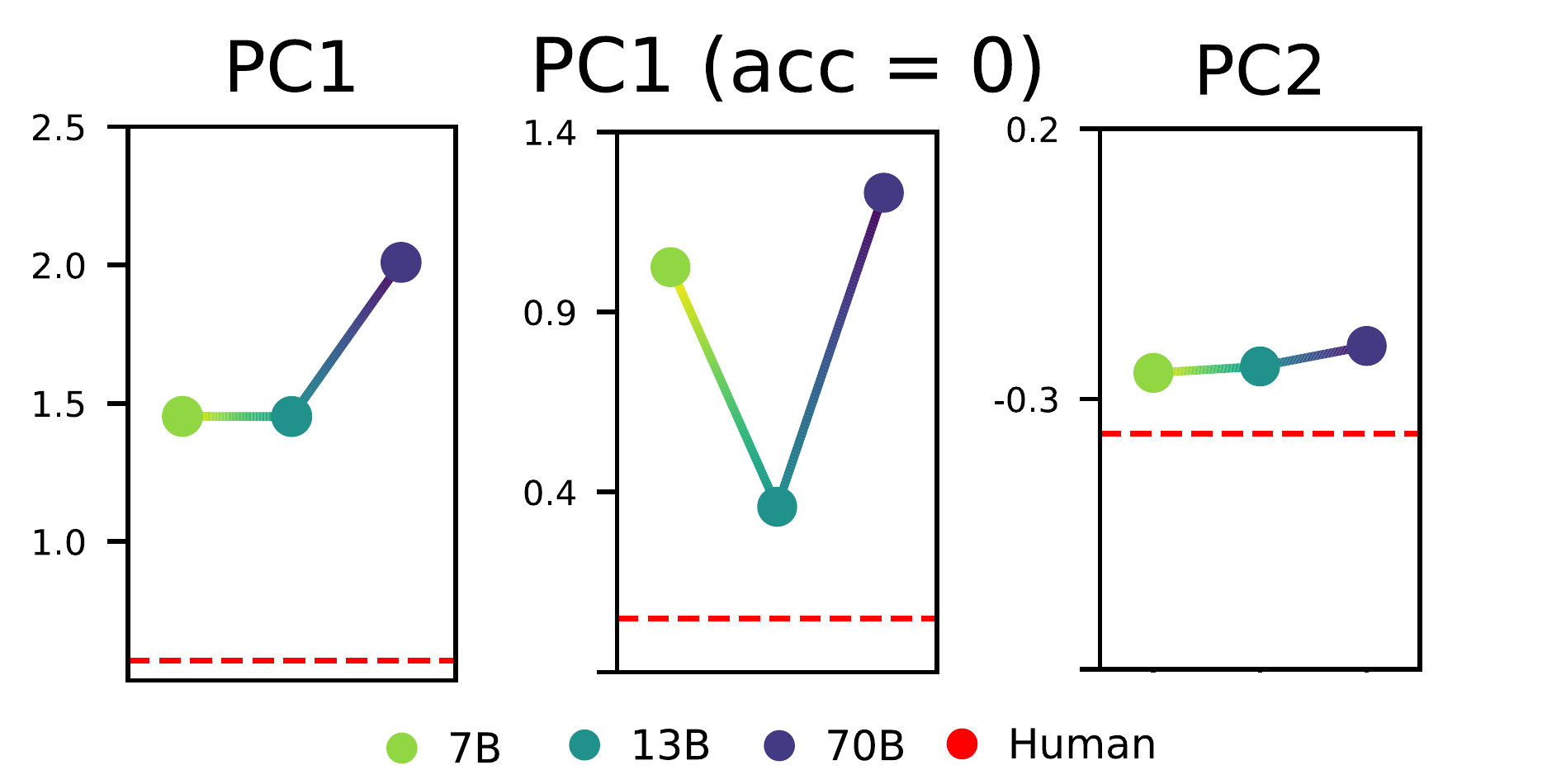
	}
	\caption{\label{fig:mreasoner-pcs_llama}}
 \end{subfigure}
 	\caption{(a) Llama 2's behavior on the 37 syllogisms whose only valid conclusion
	is ``nothing follows'' (left) and the syllogisms that license conclusions
	other than ``nothing follows'' (right). (b) Llama 2's behavior when analyzed using the principal components of the mReasoner space (see~Section~\ref{sec:mentalmodels}). We find an increased
	signature of deliberative reasoning as a function of model size, but we no
	longer observe this effect when we control for accuracy (setting the
	probability of the correct answers to zero before projecting the models' behavior
	into this space).}
\end{figure*}

We also repeated the mReasoner analysis (described in Section~\ref{sec:mentalmodels}) to analyze Llama~2 models' behavior. Although the Llama 2 models, like PaLM~2, exhibit increased
signatures of deliberative reasoning compared to mReasoner when all of their
predictions are considered (Figure~\ref{fig:mreasoner-pcs_llama}, left), when we control
for accuracy by setting the probability of the correct answer to zero for all
Llama~2 models, we find no significant correlation between model size and signatures
of deliberative reasoning (Figure~\ref{fig:mreasoner-pcs_llama}, center; cf. Figure~\ref{fig:mreasoner-pcs} for PaLM~2, where we do find such signatures even after controlling for accuracy). 

\begin{figure*}
	\centering
	\includegraphics[width=\linewidth, clip, trim=11cm 7cm 9cm 8cm]{
		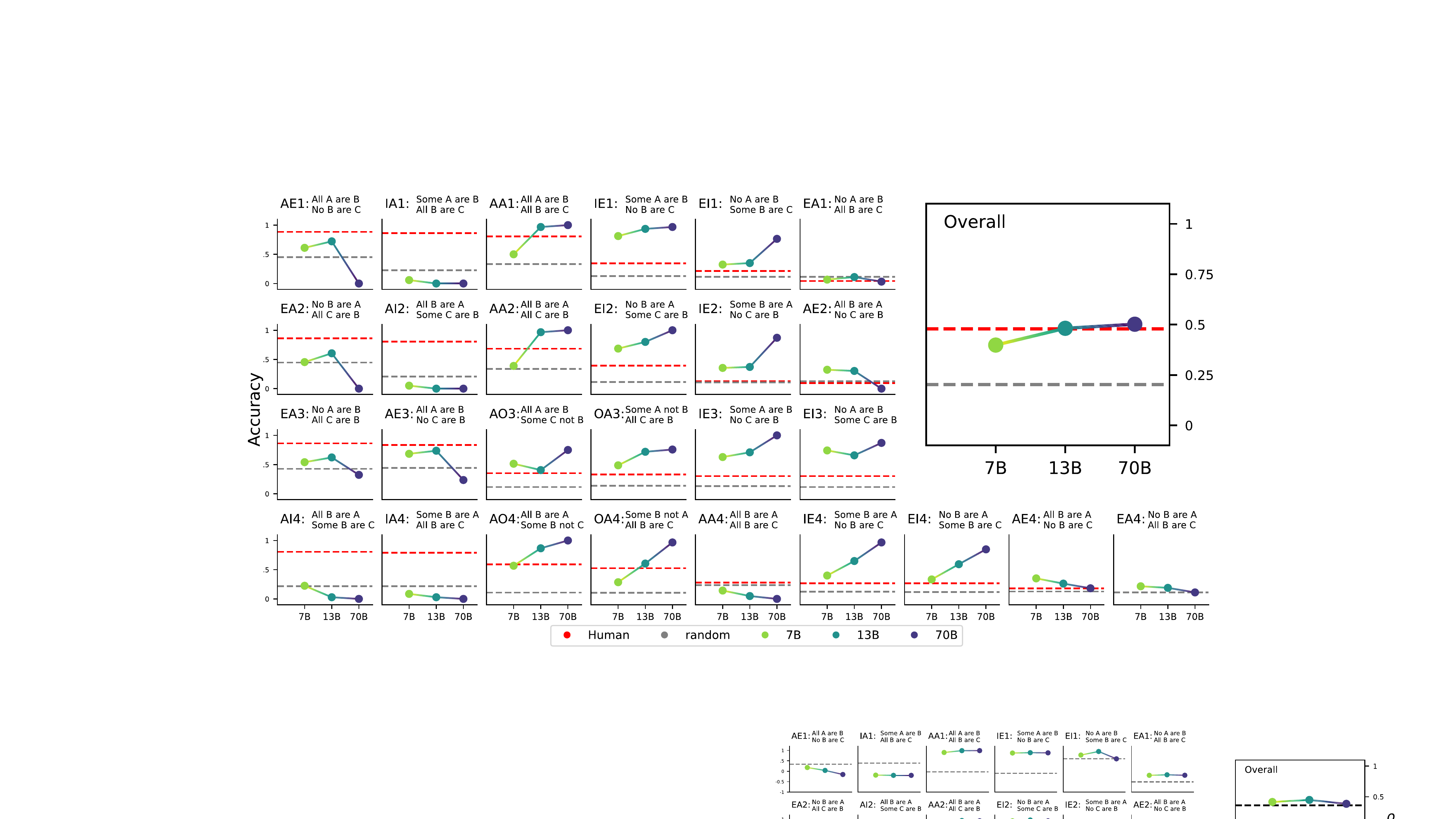
	}
	\caption{Accuracy of Llama 2 models, humans (red), and random guessing (grey).
	Random guessing accuracy differs by syllogism as some syllogisms have more than
	one valid conclusion. Syllogisms are partitioned into variable ordering (by row)
	and ordered by decreasing human accuracy from left to right. The top right inset
	shows the average accuracy across all syllogisms. Syllogisms are identified
	with the letters of the moods of the premises (Table~\ref{table:moods-variables},
	left) and the number associated with their variable ordering (Table~\ref{table:moods-variables},
	right).}
	\label{fig:hit-rate_llama}
\end{figure*}

\begin{figure*}
	\centering
	\includegraphics[width=\linewidth, clip, trim=11cm 12cm 11cm 13cm]{
		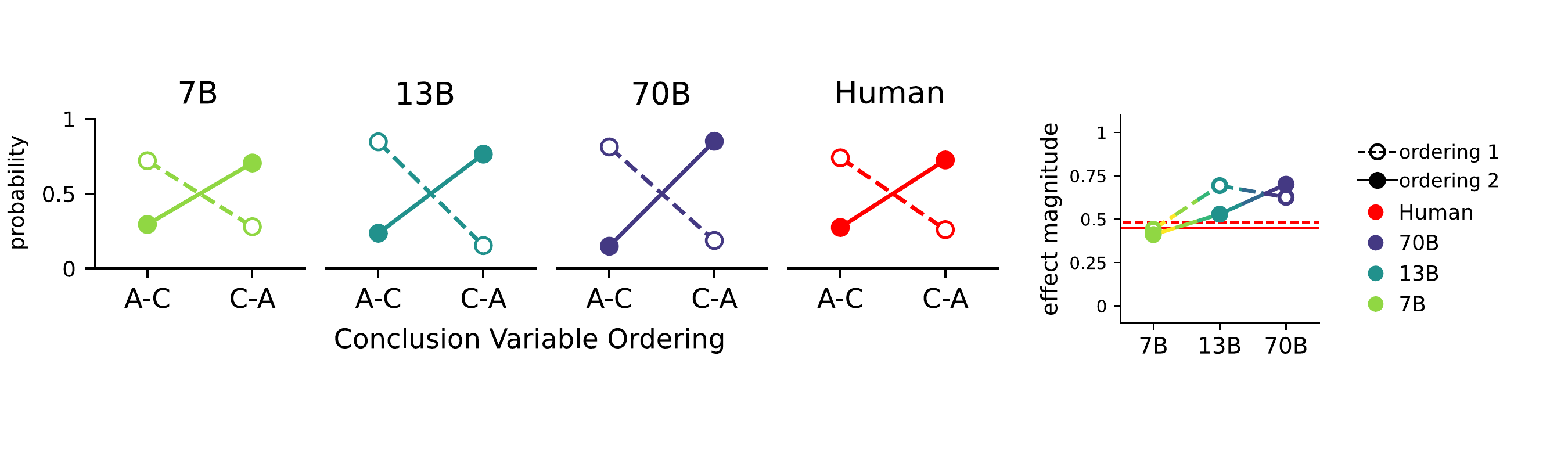
	}
	\caption{Variable ordering effects on Llama~2 responses. \textbf{Left:} The marginal probabilities of A-C and C-A ordered
	conclusions as estimated from human and LM responses. Humans and LMs both show
	variable ordering effects in the same direction. \textbf{Right:} The magnitude of
	the variable ordering effect (the absolute value of the difference between the
	probability of the C-A ordering and the probability of the A-C ordering).}
	\label{fig:figural-effect_llama}
\end{figure*}

